\documentclass{article}

\usepackage[preprint]{neurips_2026}

\usepackage{algorithm,algpseudocode}
\usepackage{algorithmicx}

\usepackage{graphicx}
\usepackage{amssymb}
\usepackage{booktabs}

\usepackage[utf8]{inputenc} 
\usepackage[T1]{fontenc}    
\usepackage{hyperref}       
\usepackage{url}            
\usepackage{booktabs}       
\usepackage{amsfonts}       
\usepackage{nicefrac}       
\usepackage{microtype}      
\usepackage{xcolor}         

\usepackage{color}
\usepackage{amsmath}
\usepackage{amssymb}
\usepackage{mathtools}
\usepackage{arydshln}
\usepackage{amsthm}
\usepackage{bm}
\usepackage{enumitem}
\usepackage{multirow}

\usepackage{tikz}

\title{Variational Test-time Optimization for\\Diffusion Synchronization}

\author{
Hyunsoo Lee\textsuperscript{\normalfont 1,2}\thanks{ Equal contribution.} ~\thanks{Work done while visiting UC Irvine.}  \hspace{5mm} 
Farrin Marouf Sofian\textsuperscript{\normalfont 2}\footnotemark[1] \hspace{5mm} 
Kushagra Pandey\textsuperscript{\normalfont 2} \hspace{5mm} 
Stephan Mandt\textsuperscript{\normalfont 2} \\ 
   \textsuperscript{1}Seoul National University \hspace{5  mm}  
   \textsuperscript{2}University of California, Irvine  \hspace{10mm} \\
   {\tt\small    philip21@snu.ac.kr, \{fmaroufs,\,pandeyk1,\,mandt\}@uci.edu}
}

\newcommand{\methodname}{SyncVC}

\usepackage{amsmath,amsfonts,bm}

\def\eqref#1{equation~\ref{#1}}

\def\1{\bm{1}}

\def\rvu{{\mathbf{i}}}

\def\rvu{{\mathbf{u}}}

\def\rvx{{\mathbf{x}}}

\def\vmu{{\bm{\mu}}}

\def\vphi{{\bm{\phi}}}

\def\vy{{\bm{y}}}

\def\mU{{\bm{U}}}

\def\mX{{\bm{X}}}

\DeclareMathAlphabet{\mathsfit}{\encodingdefault}{\sfdefault}{m}{sl}
\SetMathAlphabet{\mathsfit}{bold}{\encodingdefault}{\sfdefault}{bx}{n}

\def\gL{{\mathcal{L}}}

\newcommand{\EE}{\mathbb{E}}

\newcommand{\R}{\mathbb{R}}

\newcommand{\KL}{D_{\mathrm{KL}}}

\begin{document}

\maketitle


\begin{abstract}
Collaborative generation, which coordinates multiple diffusion trajectories to extend the capabilities of pretrained priors, has emerged as a powerful paradigm for extending the applicability of diffusion models. 
Among existing approaches, diffusion synchronization provides a scenario-agnostic solution by introducing general guidance mechanisms.
However, current synchronization approaches rely heavily on heuristics and still require task-specific tailoring, which limits their generalizability and performance.
In this work, we mathematically derive a synchronization framework based on optimal control, providing a principled explanation of diffusion synchronization.
During sampling, we optimize control variables to guide multiple trajectories toward coherent solutions while remaining close to the underlying diffusion prior. 
Our method operates entirely at test-time without additional training, thereby enabling broad applicability across diverse generation scenarios when combined with strong pretrained priors.
We demonstrate consistent improvements over baselines on three representative collaborative generation tasks, covering a wide range of modalities and applications.
Beyond performance gains, our work establishes a novel foundation for collaborative generation, opening a principled path toward extending pretrained generative models to new collaborative generation settings.
Project Website: \url{https://hleephilip.github.io/SyncVC/}.
\end{abstract}


\section{Introduction}
\label{sec:intro}

\begin{figure}[t!]
	\centering
    	\includegraphics[width=1.0\linewidth]{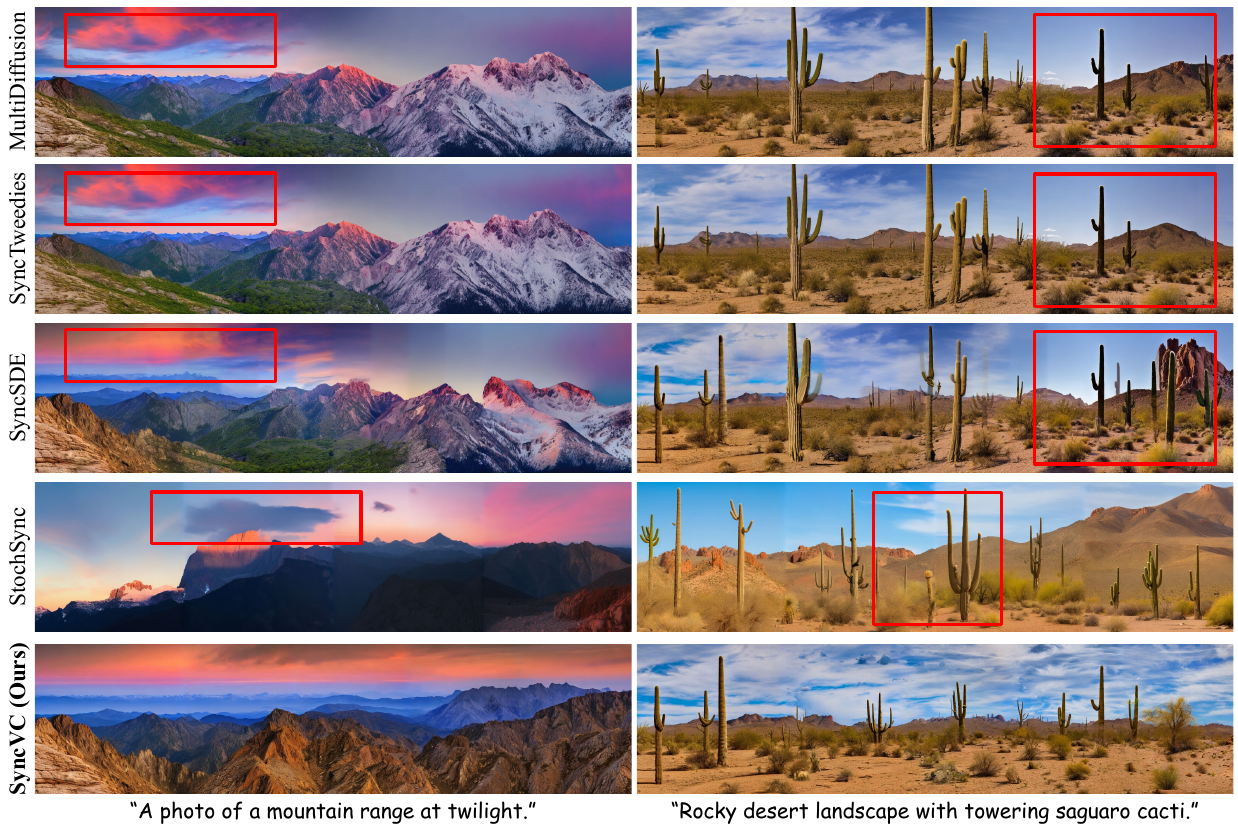}
	\caption{\textbf{Our method produces style-consistent,  high-quality wide images, outperforming all baselines.} 
    (Left) Only ours maintains a unified sky and mountain style, while baselines suffer from color inconsistency and structural discontinuities. 
    (Right) \methodname~shows consistent sky, cacti, and ground appearance, whereas the others show varying colors in the sky and cacti, along with boundary artifacts.
    }
\label{fig:wide_image_cmp}
\end{figure}

Diffusion models~\cite{sohl2015deep, ho2020denoising, pandey2023complete} and flow-matching frameworks~\cite{lipman2022flow, liu2022flow} have demonstrated strong generative priors, achieving impressive results within their training domains~\cite{rombach2022high, saharia2022photorealistic, esser2024scaling, podell2024sdxl, flux2024, xie2025sana, ho2022video, yang2023diffusionvideo, yang2024cogvideox, lee2024grid, liu2023audioldm, liu2025flashaudio, tevet2022human, karunratanakul2023guided}. 
However, extending these models beyond their native regimes, such as generating long-horizon structures from short-horizon training, still requires heavy retraining or engineering, limiting practical usability. 
This highlights the importance of \emph{collaborative generation}~\cite{lee2025syncsde}, where multiple diffusion trajectories are coupled so that they are mutually consistent while each remains plausible under the diffusion prior.
While several methods address specific tasks of collaborative generation~\cite{lee2023syncdiffusion, tang2023mvdiffusion, zhang2024taming, cai2024magic, geng2024visualanagrams, xu2025diffusion, cohan2024flexible, zhang2024texpainter, richardson2023texture, liu2023text, youwang2024paint}, many rely on task-specific heuristics, limiting generalizability and requiring substantial engineering effort to extend to new settings. 
A more desirable paradigm is \emph{diffusion synchronization}~\cite{bar2023multidiffusion, kim2024synctweedies, lee2025syncsde, yeo2025stochsync}, which provides task-agnostic and unified guidance for collaborative generation. 
Since training a diffusion model to generate multiple coordinated trajectories is computationally expensive, synchronization is performed via test-time guidance.
Rather than requiring task-specific strategies, diffusion synchronization offers a general approach that can be integrated into arbitrary priors, enabling scalable content creation across diverse scenarios.

Despite recent research, existing diffusion synchronization methods are largely driven by heuristics, such as relying on extensive empirical tests over numerous strategies~\cite{kim2024synctweedies} or introducing impractical Gaussian approximations of conditional scores~\cite{lee2025syncsde}. 
Consequently, these approaches fail to provide a principled understanding, leading to suboptimal performance and limited applicability across diverse tasks.
In this work, we address this limitation by deriving a control-based framework for diffusion synchronization. 
We introduce control variables into the diffusion process and formulate synchronization as a variational inference problem over trajectories.

At each diffusion timestep, we optimize the control variables using a novel loss function derived from our theoretical formulation. 
It balances two competing goals of collaborative generation: enforcing consistency across trajectories while remaining close to the pretrained diffusion prior. 
This formulation provides a principled explanation for collaborative generation, moving beyond heuristic approaches and interpreting it as controlled sampling. 
This principle yields a well-founded synchronization mechanism while still allowing task-specific parameterizations.
To the best of our knowledge, this work is the first to propose a unified framework for collaborative generation based on optimal control. 
We refer to our method as \textbf{Synchronized Diffusion with Variational Controls (\methodname)}.

The proposed framework is not only mathematically grounded but also widely applicable.
Since it operates through test-time optimization, it can be applied to diverse pretrained generative models~\cite{DeepFloydIF, rombach2022high, zhang2023adding, xie2025sana, le2025one} without additional training. 
Moreover, it naturally extends across diverse collaborative  generation tasks, regardless of modality. 
We validate our approach using three representative tasks: wide image generation, optical illusion generation, and 3D mesh texturing, where our method consistently outperforms baselines, as shown in Figure~\ref{fig:wide_image_cmp}.
Furthermore, unlike prior approaches, \methodname~accommodates external constraints such as style guidance without requiring redesign of an overall framework.
This flexibility highlights the advantage of our principled formulation, enabling both strong performance and practical applicability.
We summarize our contributions as follows:
\begin{itemize}
    \item We propose \methodname, a mathematically grounded test-time optimization framework for collaborative generation, providing a fundamental explanation of diffusion synchronization.
    \item \methodname~introduces control-based guidance for generative modeling, where control variables are optimized via a variational objective, yielding a general and extensible sampling mechanism across tasks and modalities.
    \item Our method incurs no training cost and is broadly applicable, enabling direct integration with pretrained diffusion priors while naturally benefiting from advances in stronger models.
    \item We demonstrate strong empirical performance across diverse collaborative generation tasks, spanning both 2D and 3D generation scenarios.
\end{itemize}


\section{Related work}
\label{sec:rel_work}

\textbf{Task-specific methods for collaborative generation.} A representative example of collaborative generation is wide image generation, where multiple trajectories for fixed-size, partially-overlapping patches are fused into a single wide image. 
SyncDiffusion~\cite{lee2023syncdiffusion} ensures consistent style along the wide image by minimizing LPIPS distance~\cite{zhang2018unreasonable} between patches.
Another task is optical illusion (ambiguous image) generation, which synthesizes a single image that conveys different semantics under different transformations.
Anagram-MTL~\cite{xu2025diffusion} formulates this task as a  multi-task learning problem~\cite{zhang2018overview, zhang2021survey} with attention-based regularization and CLIP-based~\cite{radford2021learning} adaptive noise reweighting.
In 3D graphics, text-guided mesh texturing requires consistency across multiple views and has been addressed using diffusion-based approaches~\cite{zhang2024texpainter, youwang2024paint, richardson2023texture, liu2023text, zeng2024paint3d, chen2023text2tex, yan2024flexipainter}. 
For example, TexPainter~\cite{zhang2024texpainter} enforces multi-view consistency via color-space fusion at each diffusion step, guiding the denoising process for coherent texture generation.
Meanwhile, TEXTure~\cite{richardson2023texture} employs a texturing-tailored diffusion process with a trimap representation, and iteratively updates texture maps from different viewpoints. 
However, these methods are tailored to specific tasks and lack generalizability across different collaboration scenarios. 
In contrast, our method does not rely on task-specific designs, but provides a general framework applicable to arbitrary tasks and modalities with strong performance.

\textbf{Diffusion synchronization.} Diffusion synchronization methods~\cite{bar2023multidiffusion, kim2024synctweedies, lee2025syncsde, yeo2025stochsync} aim to provide general mechanisms across diverse collaborative generation scenarios. 
MultiDiffusion~\cite{bar2023multidiffusion} aligns trajectories by optimizing diffusion latents with respect to a heuristically designed objective, resulting in a closed-form solution via latent averaging. 
SyncTweedies~\cite{kim2024synctweedies} empirically evaluates 60 synchronization strategies and selects the best-performing configuration, identifying averaged Tweedie estimates~\cite{robbins1992empirical} as the optimal strategy.
However, its reliance on heuristics limits its scalability and generalization beyond the evaluated settings.
On the other hand, SyncSDE~\cite{lee2025syncsde} proposes auto-regressive trajectory sampling, where the conditional score of the current trajectory given the previously generated trajectory is approximated with a Gaussian. 
This strong assumption limits its general applicability. 
StochSync~\cite{yeo2025stochsync} builds upon the SyncTweedies and interprets synchronization via score distillation sampling (SDS)~\cite{poole2022dreamfusion}, but still depends on heuristic engineering techniques such as non-overlapping view sampling.
In contrast, our method introduces control variables into the sampling process and optimizes them using a loss function derived from variational inference. 
This minimizes the need for heuristic modeling, providing a principled framework that leads to improved performance across a wide range of collaborative generation tasks.

\textbf{Diffusion with optimal controls.} Recently, optimal control~\cite{kappen2008stochastic} has been used to design guidance in diffusion models~\cite{huang2024symbolic, pandey2025variational, geyfman2026calibrated, li2024solving, berner2024an, azangulov2025adaptive, chen2024generative, rout2025rbmodulation}.
Stochastic Control Guidance~\cite{huang2024symbolic} formulates guidance as a stochastic optimal control problem, leveraging path-integral control for plug-and-play guidance with non-differentiable rewards.
For general inverse problems, Diffusion Trajectory Matching~\cite{pandey2025variational} formulates guidance as a variational control problem, where control variables are optimized to follow terminal constraints while regularizing deviations from the pretrained diffusion prior; related fast samplers have also been studied for iterative refinement models~\cite{pandey2024fast}. 
\textit{Azangulov et al.}~\cite{azangulov2025adaptive} formulates adaptive guidance scheduling as a stochastic optimal control problem, dynamically selecting the guidance scale via controls.
While these approaches primarily focus on guiding a single diffusion trajectory with external constraints, our work addresses \emph{collaborative generation}, where multiple trajectories must be coordinated simultaneously. 
To the best of our knowledge, we are the first to introduce a control-based framework for diffusion synchronization.


%
\begin{figure}[t!]
	\centering
    	\includegraphics[width=1.0\linewidth]{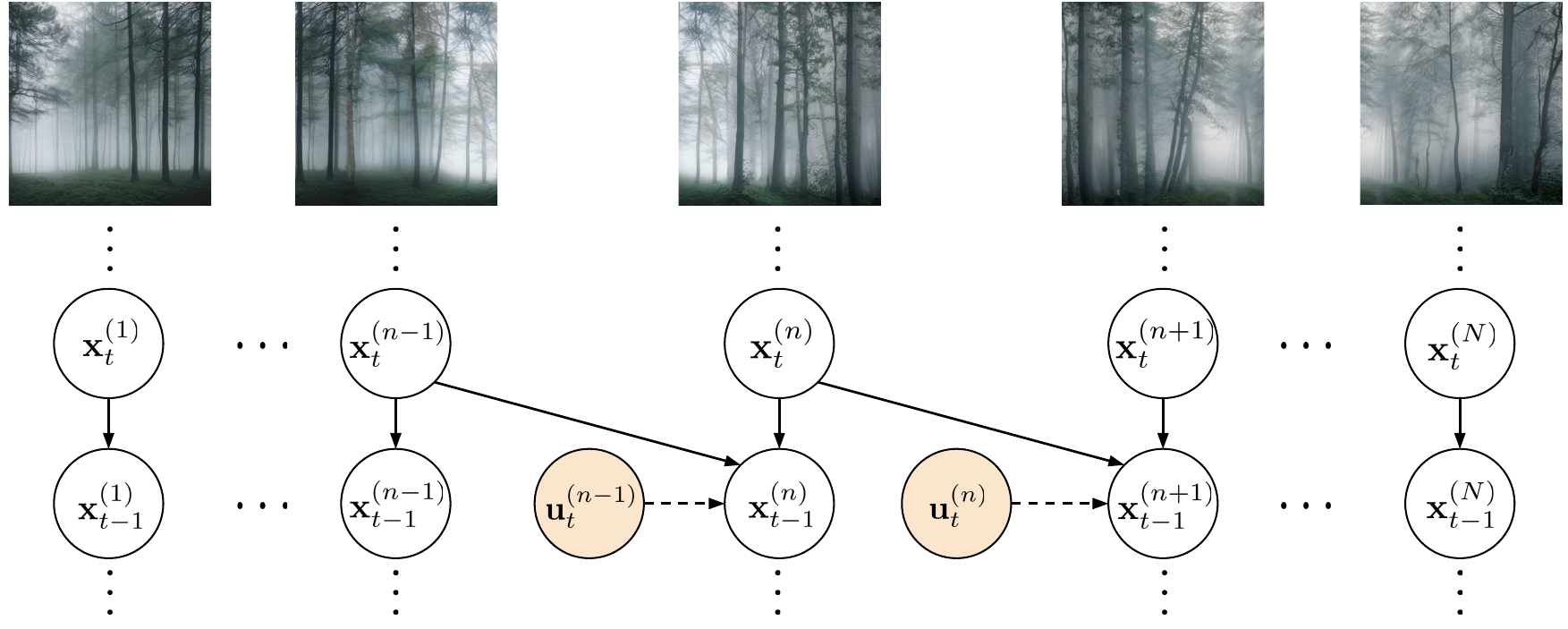}
	\vspace{-6mm}
	\caption{
    \textbf{Overall mechanism of \methodname.}
    Control variables are introduced into the diffusion process for collaborative generation through synchronized diffusion.
    We visualize the case of wide image generation, where each diffusion trajectory models a partially overlapping image patch.
    }
    \vspace{-1mm}
\label{fig:method}
\end{figure} 

\section{Collaborative Generation with Synchronized Variational Controls}
\label{sec:method}

\paragraph{Problem formulation.}
Given an observation $\vy \in \R^m$, our goal is to generate a set of $N$ consistent elements $\mX = \{\rvx_0^{(1)}, \ldots,  \rvx_0^{(N)}\}$ ($\rvx_0^{(i)} \in \R^d$) that maximizes the likelihood of the observation $\vy$. For instance, for wide image generation, the sequence elements can be overlapping image patches. We begin by defining the likelihood $p(\vy \mid \mX)$ as an energy-based reward function, $r: \R^m \times \R^{Nd} \rightarrow \R$,
\begin{equation}
    p(\vy \mid \mX) \propto \exp{(r(\vy, \mX))}.
    \label{eq:likelihood_def_via_energy}
\end{equation}
The form of the reward is task-specific. For instance, for stylized wide image generation, the reward can be defined as maximizing the overlap between consecutive sequence elements~\cite{lee2025syncsde}, and $\vy$ may additionally encode conditioning information from a style transfer task~\cite{gatys2016image}. 
We will discuss several parameterizations of the rewards considered in this work in the following paragraphs.
For the remainder of this work, we restrict our focus to rewards which are \textbf{known}, \textbf{differentiable}, and can be evaluated in \textbf{closed form}.
We model the prior over each set element $p(\rvx_0^{(n)})$ with a pretrained diffusion model using $T$ denoising steps,
\begin{equation}
    p(\rvx_0^{(n)}) = \int p(\rvx_T^{(n)}) \prod_t p_\vphi(\rvx_{t-1}^{(n)} \mid \rvx_{t}^{(n)})\, d\rvx_{1:T}^{(n)},
    \label{eq:prior}
\end{equation}
where $p_\vphi$ denotes the denoising kernel with mean $\vmu_\vphi(\rvx_t^{(n)}, t),$ and $t$ denotes the diffusion timestep. 
The distribution $p(\rvx_T^{(n)})$ is typically modeled as a standard Gaussian. We make two further assumptions.
First, we operate under the regime of training-free test-time guidance and thus keep the pretrained diffusion model fixed.
Secondly, while the pretrained diffusion model can also be conditioned on additional information (\textit{e.g.} a text prompt~\cite{rombach2022high, xie2025sana}), we drop this in the notation for convenience. 
As noted earlier, we aim to generate a sequence $\mX$ which maximizes the likelihood of $\vy$. 
Formally, we want to sample from the following tilted distribution,
\begin{equation}
    q(\mX \mid \vy) \propto p(\mX)\exp{(r(\vy, \mX) / \beta)}.
    \label{eq:tilted_distribution}
\end{equation}
We define a prior over the sequence as $p(\mX) = \prod_n^N p(\rvx_0^{(n)})$. We use this factorized prior for simplicity and because it works well empirically (see Sec.~\ref{sec:experiments}). However, we note that this is not a limitation of our framework, as more expressive parameterizations of the prior are possible and can be an interesting direction for future work. 
In most cases, this tilted distribution $q(\mX \mid \vy)$ is intractable to compute in closed form. 
Therefore, we rely on variational inference to approximate the latter, which we discuss next.

\paragraph{Diffusion synchronization via variational inference.}
We define the variational distribution over the sequence $\mX$ with a diffusion process:
\begin{equation}
    q(\mX \mid \vy) = \int q\!\left(\rvx_{{0:T}}^{(1)}\right) \prod_{n=2}^{N} \prod_{t=1}^{T} q\!\left(\rvx_{t-1}^{(n)} \mid \rvx_t^{(n)}, \rvx_t^{(n-1)}, \vy\right) d\rvx_{1:T}^{(1:n)}.
    \label{eq:variational_q}
\end{equation}
This factorization is natural for an ordered sequence $\{\rvx_0^{(1)}, \ldots, \rvx_0^{(N)}\}$ and may be sub-optimal in settings without such ordering, but we find that it works well empirically (see Sec.~\ref{sec:experiments}) and therefore do not explore alternative schemes. 
By conditioning the reverse transition $q(\rvx_{t-1}^{(n)} \mid \rvx_t^{(n)}, \rvx_t^{(n-1)}, \vy)$ on the noisy latent $\rvx_t^{(n-1)}$ of the previous trajectory, the generation of $\rvx_0^{(n)}$ is synchronized with $\rvx_0^{(n-1)}$ at every diffusion step.

A natural choice to model the distribution $q(\rvx_{t-1}^{(n)} \mid \rvx_t^{(n)}, \rvx_t^{(n-1)}, \vy)$ is a conditional Gaussian approximation~\cite{lee2025syncsde}. To steer generation toward the reward $r$, we augment the variational distribution with additional variational parameters $\rvu_t^{(n-1)}$ at each step.
These auxiliary variables couple adjacent denoising trajectories, thereby enabling collaborative generation (see Figure~\ref{fig:wide_image_small_overlap}). Combining the generative model in Eq.~\ref{eq:prior} with the augmented variational distribution, the evidence lower bound (ELBO) takes the form (see Appendix~\ref{sec:supp_method}),
\begin{align}
    \gL(\vy) :=\; \EE_{q}[r\!\left(\vy, \mX\right)]
    - \lambda \sum_{n=2}^{N} \sum_{t=1}^{T} \! \KL\!\left( q\!\left(\rvx_{t-1}^{(n)} \mid \rvx_t^{(n)}, \rvx_t^{(n-1)}, \rvu_t^{(n-1)}, \vy\right) \,\|\, p\!\left(\rvx_{t-1}^{(n)} \mid \rvx_t^{(n)}\right) \right).
    \label{eq:elbo_simplified}
\end{align}
Eq.~\ref{eq:elbo_simplified} captures the two competing goals of collaborative generation. The first term encourages sequences that maximize the expected reward, while the second pulls samples from the variational distribution toward the noisy submanifold defined by the prior diffusion model at each denoising step. The scalar hyperparameter $\lambda$ controls the strength of this regularization. For any observation $\vy$, we optimize Eq.~\ref{eq:elbo_simplified} to infer the variational parameters $\rvu_t^{(n-1)}$ directly at test time. We parameterize the augmented variational distribution as a unimodal Gaussian distribution with mean
\begin{align}
    \bar{\vmu}_t^{(n)} = \underbrace{\vmu_\vphi\!\left(\bar{\rvx}_t^{(n)}, t\right)}_{\text{Terminal Step}} - \frac{\gamma}{2} \sigma_t^2 \, \underbrace{\nabla_{\rvx_t^{(n)}} \left\| f \left( \rvx_t^{(n-1)}, \vy  \right) - \bar{\rvx}_t^{(n)} \right\|_2^2}_{\text{Regularizer}},
    \label{eq:q_x_param}
\end{align}
where $\bar{\rvx}_t^{(n)} = \rvx_t^{(n)} + \beta \rvu_t^{(n-1)}$ and $\beta > 0$ is a hyperparameter that defines the strength of the controls. 
$f(\cdot, \cdot)$ is an operator defined by the reward function, and practical choices are described in the next paragraph. The first term perturbs the denoising trajectory at time $t$ along the direction of $\rvu_t^{(n-1)}$ to maximize the reward; following~\citep{pandey2025variational}, we refer to these auxiliary variables as \emph{variational controls}. The second term steers the trajectory to reduce the gap between adjacent denoising chains, acting as a regularizer. Together, the two terms guide each denoising trajectory toward higher reward while maintaining consistency across trajectories. We refer to this overall framework as \textbf{Synchronized Diffusion with Variational Controls (\methodname)} and summarize it in Fig.~\ref{fig:method}.

\paragraph{Choice of the reward.}
Because the variational distribution in Eq.~\ref{eq:variational_q} generates the sequence autoregressively, the reward function must respect the same causal ordering. We therefore decompose the overall reward into a sum of sub-rewards, where the $n$-th term depends only on the current element and those preceding it:
\begin{equation}
    r\!\left(\vy, \mX \right) := \sum_{n=2}^{N} \tilde{r}\!\left(\vy, \rvx_0^{(1)}, \dots, \rvx_0^{(n)}\right).
    \label{eq:reward_factorization}
\end{equation}
This decomposition exposes a natural design principle: each $\tilde{r}$ should measure how well the new element $\rvx_0^{(n)}$ agrees with the elements already generated, under a task-specific notion of consistency. 
Below, we instantiate $\tilde{r}$ for the three collaborative generation tasks studied in this work. We deliberately keep these designs simple and intuitive rather than relying on heavily engineered components; as shown in Sec.~\ref{sec:experiments}, even these straightforward choices already outperform prior baselines, and we leave more sophisticated parameterizations to future work.

\textit{Wide image generation.}
The goal is to synthesize a horizontally elongated image from a text prompt $\vy$, where each sequence element $\rvx_0^{(n)}$ corresponds to a patch and adjacent patches partially overlap. Consistency then amounts to agreement on the shared region between neighbors:
\begin{equation}
    \tilde{r}\!\left(\vy, \rvx_0^{(1)}, \dots, \rvx_0^{(n)}\right) = - \frac{\gamma}{2} \left\| \mathbf{M} \odot \left( f\!\left(\rvx_0^{(n-1)}\right) - \rvx_0^{(n)} \right) \right\|^2_2,
    \label{eq:wide_image_reward}
\end{equation}
where $f(\cdot)$ shifts its input along the $x$-axis by the patch stride, $\mathbf{M}$ is a binary mask selecting the overlap region, and $\gamma$ is a tunable weight. Intuitively, this reward pulls the left side of the $n$-th patch toward the right side of the $(n{-}1)$-th patch.
Although our framework can incorporate more general reward functions, we use Eq.~\ref{eq:wide_image_reward} as the main reward because it yielded the best results in our experiments.
We also evaluate a CLIP-augmented variant~\cite{radford2021learning} that adds a semantic guidance term based on the similarity between $\rvx_0^{(n)}$ and the text prompt $\vy$, and provide results in Appendix~\ref{sec:supp_exp_details}.

\textit{Optical illusion generation.}
The task is to synthesize a single image whose semantic content changes under a fixed transformation, such as rotation or flipping. The observation $\vy$ comprises two text prompts, and we sample two trajectories---one per prompt---under the corresponding views. Consistency here means that the two trajectories should agree once the transformation is applied:
\begin{equation}
    \tilde{r}\!\left(\vy, \rvx_0^{(1)}, \rvx_0^{(2)}\right) = - \frac{\gamma}{2} \left\| f\!\left(\rvx_0^{(1)}\right) - \rvx_0^{(2)} \right\|^2_2,
    \label{eq:illusion_reward}
\end{equation}
where $f(\cdot)$ is the illusion transformation operator. The reward therefore encourages the second trajectory to match the transformed version of the first, so that both prompts are simultaneously satisfied in a single image.

\textit{Text-guided 3D mesh texturing.}
The main challenge in this task is multi-view consistency: each trajectory $\rvx_0^{(n)}$ generates a 2D image from a different viewpoint, and these images must collectively define a coherent texture on the input mesh. Here $\vy$ consists of the text prompt together with the source mesh. To define the sub-reward at step $n$, we first bake an auxiliary texture from the previously generated views $\{\rvx_0^{(j)}\}_{j=1}^{n-1}$, then render this texture from the $n$-th viewpoint and compare with $\rvx_0^{(n)}$:
\begin{equation}
    \tilde{r}\!\left(\vy, \rvx_0^{(1)}, \dots, \rvx_0^{(n)}\right) = - \frac{\gamma}{2} \left\| \mathbf{M}^{(n)} \odot \left( f\!\left(\rvx_0^{(1)}, \dots, \rvx_0^{(n-1)}, \vy, n\right) - \rvx_0^{(n)} \right) \right\|^2_2,
    \label{eq:mesh_texturing_reward}
\end{equation}
Here, $f(\cdots, \vy, n)$ composes texture baking and rendering: texture baking fuses the previous multi-view images into a texture map, and rendering projects that texture from the $n$-th viewpoint. The mask $\mathbf{M}^{(n)}$ selects the foreground region in the rendered image. In effect, this reward asks each new view to remain faithful to what the texture already implies from earlier views.

\vspace{-1mm}
\paragraph{Practical considerations.}
The reward functions in Eqs.~\ref{eq:wide_image_reward} and~\ref{eq:illusion_reward} are defined on the clean sequence elements $\rvx_0^{(n)}$. Direct optimization of the variational objective in Eq.~\ref{eq:elbo_simplified} would therefore require rolling out the full reverse diffusion chain at every timestep to evaluate $r(\vy, \mX)$, which is computationally prohibitive. Instead, we approximate the clean sample at the current timestep using Tweedie's estimate. Therefore, the loss function simplifies to the following objective:
\begin{equation}    
    \rvu_t^{\star} = \arg\min_{\rvu_t} \;
    \sum_{n=2}^{N}  \left[ - \tilde{r}\!\left(\vy, \hat{\rvx}_{0|t}^{(1)}, \dots, \hat{\rvx}_{0|t}^{(n)}\right)
    + \lambda \left\| \bar{\vmu}_t^{(n)} - \vmu_\vphi\!\left(\rvx_t^{(n)}, t\right) \right\|_2^2\right],
    \label{eq:practical_objective_gaussian}
\end{equation}
where Tweedie's estimate is given by $\hat{\rvx}_{0|t}^{(n)} =  \EE \left[{\rvx}_0 \mid \bar{\rvx}_t^{(n)}\right]$.

\paragraph{Reformulated objective for DDIM}.
Under the DDIM~\cite{song2020denoising} parameterization, we derive a simplified objective as follows (see Appendix~\ref{sec:supp_ddim_derivation}):
\begin{align}
    \rvu_t^{\star}
    = &\arg\min_{\rvu_t} \; \sum_{n=2}^{N}
    \left[
    - \tilde{r}\!\left(\vy, \hat{\rvx}_{0|t}^{(1)}, \dots, \hat{\rvx}_{0|t}^{(n)}\right)
    + \lambda  a_t^2 \left\| \rvu_t^{(n-1)}\right\|^2_2 \right. \notag\\
    &\left.
    + \lambda b_t^2 \left\|
    \epsilon_{\theta}(\bar{\rvx}_t^{(n)} , t)
    + \frac{\gamma}{2} \sqrt{1-\alpha_t}
    \nabla_{\rvx_t^{(n)}} 
    \left\| f \left( \rvx_t^{(n-1)}, \vy  \right)
    - \bar{\rvx}_t^{(n)}
    \right\|_2^2
    - \epsilon_{\theta}(\rvx_t^{(n)}, t)
    \right\|^2_2
    \right],
    \label{eq:practical_objective_ddim}
\end{align}
where $\epsilon_{\theta}(\cdot, \cdot)$ denotes the noise prediction network, 
\begin{equation}
        a_t = \sqrt{\frac{\alpha_{t-1}}{\alpha_t}} \quad \text{and} \quad b_t = \sqrt{1-\alpha_{t-1} }  - \sqrt{\frac{(1-\alpha_t)\alpha_{t-1}}{\alpha_t}}.
\end{equation}
%


\section{Experiments}
\label{sec:experiments}

In this section, we evaluate the practical effectiveness of \methodname~across key tasks introduced in Sec.~\ref{sec:method}. 
Our method is implemented with PyTorch~\cite{paszke2019pytorch}, and control variables are optimized using Adam optimizer~\cite{kingma2014adam}. 
DDIM~\cite{song2020denoising} and classifier-free guidance~\cite{ho2021classifierfree} are used for diffusion sampling across all tasks. 
For all tables, we \textbf{bold the best} and \underline{underline the second-best} results.
Task-specific experimental details and results are provided in the corresponding subsections and Appendix~\ref{sec:supp_exp_details}.

\subsection{Wide image generation}

\paragraph{Evaluation protocol.}

We generate $2048 \times 512$ wide images using the pretrained Stable Diffusion~\cite{rombach2022high}, with each patch of size $512^2$.
For \methodname, five patches are sampled with an overlap of 128 pixels and are sequentially composited to form the final wide image, whereas baselines use their default overlapping configurations.
We compare the proposed approach with diffusion synchronization methods~\cite{bar2023multidiffusion, lee2025syncsde, kim2024synctweedies, yeo2025stochsync}. 
For evaluation, we adopt 15 text prompts from prior works~\cite{bar2023multidiffusion, lee2025syncsde, kim2024synctweedies, yeo2025stochsync} and generate 50 wide images per prompt. 

For wide image generation, maintaining consistency across patches is the most important criterion. 
To evaluate coherence, we crop each generated wide image into four non-overlapping views and measure all possible pairwise relationships among them. 
Specifically, for perceptual and stylistic alignment, we leverage Intra-LPIPS and Intra-Style-Loss from prior work~\cite{lee2023syncdiffusion}. 
In addition, to assess color alignment, we compute color histograms in the HSV space for each non-overlapping view and measure their $\chi^2$ distance and histogram intersection.
We also measure KID~\cite{binkowski2018demystifying} using randomly cropped views, to assess image quality and diversity.

\paragraph{Results.}

Table~\ref{tab:wide_image} shows that our method consistently outperforms all baselines, with particularly large gains in Intra-Style-Loss and $\chi^2$-Histogram distance, which measure style and color consistency, respectively.
It also demonstrates strong distributional alignment as reflected by KID.
{Figure~\ref{fig:wide_image_cmp}} visualizes qualitative results. 
While baselines exhibit inconsistent styles and discontinuities along the horizontal axis, \methodname~produces smooth transitions with a unified style.
We further demonstrate that our method can be applied beyond Stable Diffusion by synthesizing high-resolution wide images using the pretrained SANA model~\cite{xie2025sana}, which provides stronger generative priors. 
These results are visualized in Appendix~\ref{sec:supp_exp_details}.

\begin{table}[t!]
\caption{
Quantitative evaluation on wide image generation.
The proposed method outperforms all baselines, with a particularly large margin in Intra-Style-Loss~\cite{gatys2016image} and $\chi^2$-Histogram distance, which measure style and color consistency across the wide image, respectively.
KID~\cite{binkowski2018demystifying} is scaled by $10^3$.
}
\vspace{-2mm}
\centering
\setlength{\tabcolsep}{2.5pt} 
\scalebox{0.85}{
    \begin{tabular}{l c c c c c}
    \toprule
    Method & MultiDiffusion~\cite{bar2023multidiffusion}  & SyncTweedies~\cite{kim2024synctweedies} & SyncSDE~\cite{lee2025syncsde}  & StochSync~\cite{yeo2025stochsync} & \textbf{\methodname~(Ours)}   \\
    \midrule
    Intra-LPIPS~\cite{zhang2018unreasonable} $\downarrow$ & 0.637 & 0.620 & 0.653 & \underline{0.617} & \textbf{0.592} \\
    Intra-Style-Loss~\cite{gatys2016image} $\downarrow$  &  \underline{58.46} & 78.05 & 63.98 & 67.56 & \textbf{44.34}  \\
    $\chi^2$-Histogram dist. $\downarrow$ &  \underline{1.211}  & 1.345  & 1.307 & 1.352 & \textbf{0.751} \\
    Histogram intersect. $\uparrow$ & \underline{0.549}  & 0.519  & 0.526  & 0.518 & \textbf{0.665} \\
    KID~\cite{binkowski2018demystifying} $\downarrow$ & 58.26 & 60.81 & \underline{57.08} & 100.47 & \textbf{52.07}  \\
    \bottomrule
    \end{tabular}
    }
    \vspace{-5mm}
    \label{tab:wide_image}
\end{table} 

\begin{figure}[h!]
	\centering
    \vspace{-3mm}
    \includegraphics[width=1.0\linewidth]{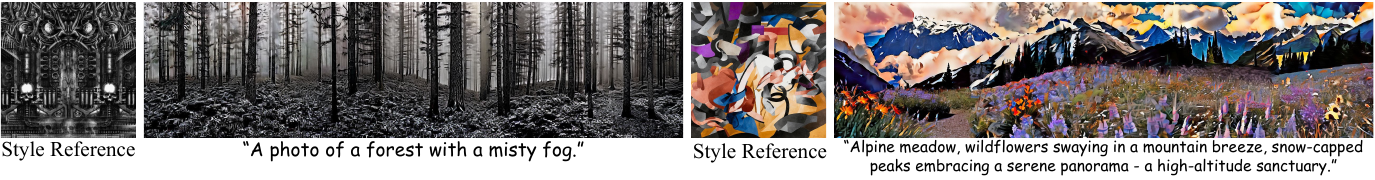}
	\vspace{-7mm}
	\caption{
    \textbf{\methodname~enables flexible generation under external constraints such as style guidance}, transferring texture and overall color from the style reference while preserving the semantics of the given prompt without artifacts.
    }
    \vspace{-3mm}
\label{fig:wide_image_misc}
\end{figure}

\paragraph{Incorporating additional conditions.}
Unlike prior works that rely on closed-form guidance, \methodname~naturally accommodates additional constraints through a reasonable reward design.
As a representative example, we consider style guidance for wide image generation by adding a style transfer loss~\cite{gatys2016image} between a style reference and $\rvx_0^{(n)}$ within the reward function of Eq.~\ref{eq:wide_image_reward}.
{Figure~\ref{fig:wide_image_misc}} (b) shows that our method can flexibly incorporate style guidance, highlighting its strength as a fundamental framework that can accommodate a broad class of reward parameterizations.

\subsection{Optical illusion generation}

\paragraph{Evaluation protocol.}
We generate images using the pretrained DeepFloyd-IF~\cite{DeepFloydIF}. 
For evaluation, we adopt 5 pairs of (transformation, prompt) from prior work~\cite{geng2024visualanagrams, lee2025syncsde} and generate 100 images for each pair. 
The final output consists of two views: the image from the second trajectory (view 2) and its inverse-transformed counterpart (view 1), both of which are used for evaluation.
We compare against synchronization methods~\cite{kim2024synctweedies, lee2025syncsde}, and a task-specific method~\cite{xu2025diffusion}.
For metrics, we measure FID~\cite{heusel2017gans}, KID~\cite{binkowski2018demystifying} to quantify distributional alignment and MUSIQ~\cite{ke2021musiq} to assess image quality.

\paragraph{Results.}

\begin{figure}[h!]
	\centering
    	\includegraphics[width=1.0\linewidth]{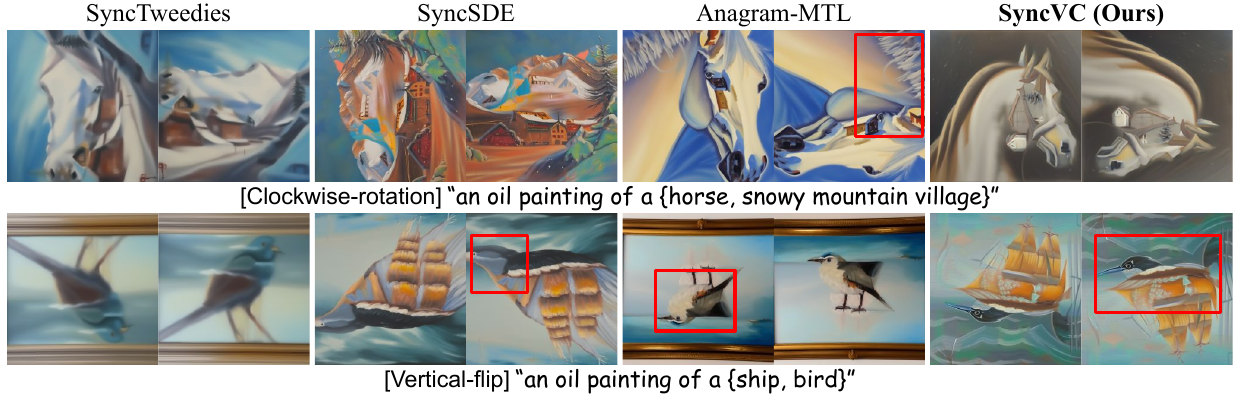}
	\vspace{-6mm}
	\caption{
    \textbf{Our method outperforms all baselines by clearly encoding both semantics under illusion while maintaining high quality.}
    For each method, we visualize both views (view 1 \& 2) of the final result. 
    (Row 1) SyncTweedies~\cite{kim2024synctweedies} produces a blurry image with low quality, Anagram-MTL~\cite{xu2025diffusion}, although tailored for this task, also generates some artifacts (denoted as bounding box).
    (Row 2) SyncTweedies still results in a blurry image, while both SyncSDE~\cite{lee2025syncsde} and Anagram-MTL struggle to simultaneously encode both semantics (bird and ship, respectively).
    }
    
\label{fig:illusion_cmp}
\end{figure}

We show quantitative results in Table~\ref{tab:illusion}, with qualitative comparisons in {Figure~\ref{fig:illusion_cmp}}. 
Our method consistently achieves outstanding performance across all metrics.
In particular, SyncTweedies~\cite{kim2024synctweedies} tends to produce blurry images with lower aesthetic scores, whereas the proposed method maintains high visual quality while clearly encoding both semantic interpretations within a single image.
This highlights both the limitations of heuristic-based modeling and the advantages of our principled formulation that coherently models the interaction between trajectories.

\begin{table}[h!]
\vspace{-2mm}
\caption{
Quantitative evaluation on optical illusion generation. 
Our approach outperforms all baselines, in terms of both distributional alignment (FID~\cite{heusel2017gans}, KID~\cite{binkowski2018demystifying}) and image quality (MUSIQ~\cite{ke2021musiq}). 
KID score is scaled by $10^3$.
}
\vspace{-2mm}
\centering
\setlength{\tabcolsep}{8pt} 
\scalebox{0.95}{
    \begin{tabular}{l c c c c}
    \toprule
    Method & SyncTweedies~\cite{kim2024synctweedies} & SyncSDE~\cite{lee2025syncsde}  & Anagram-MTL~\cite{xu2025diffusion} & \textbf{\methodname~(Ours)}   \\
    \midrule
    FID~\cite{heusel2017gans} $\downarrow$ & \underline{255.37} & 264.10 & 256.96 & \textbf{252.87} \\
    KID~\cite{binkowski2018demystifying} $\downarrow$ & 195.52 & \underline{176.14} & 190.64 & \textbf{175.47}  \\
    MUSIQ~\cite{ke2021musiq} $\uparrow$ & 32.28 & \underline{52.74} & 41.92 & \textbf{54.40}  \\
    \bottomrule
    \end{tabular}
    }
    \vspace{-3mm}
    \label{tab:illusion}
\end{table}

\subsection{Text-guided 3D mesh texturing}

\paragraph{Evaluation protocol.}

We generate diffusion trajectories using the pretrained depth-conditioned ControlNet~\cite{zhang2023adding}. 
For \methodname, we define 8 trajectories by uniformly sampling azimuth angles at a fixed elevation, producing partially overlapping views, while following default setups for baselines. 
Images generated from each trajectory are used to synthesize the final texture.
We compare our method against synchronization approaches~\cite{kim2024synctweedies, lee2025syncsde, yeo2025stochsync} and task-specific methods~\cite{zhang2024texpainter, richardson2023texture}.
For evaluation, we use 350 (mesh, prompt) pairs sampled from the Objaverse dataset~\cite{objaverse}. 
Each textured mesh is rendered from 10 viewpoints, and the resulting images are used to compute FID~\cite{heusel2017gans}, KID~\cite{binkowski2018demystifying}, and CLIP image–text similarity (CLIP-S)~\cite{radford2021learning}.

\paragraph{Results.}

\begin{figure}[h!]
	\centering
    	\includegraphics[width=1.0\linewidth]{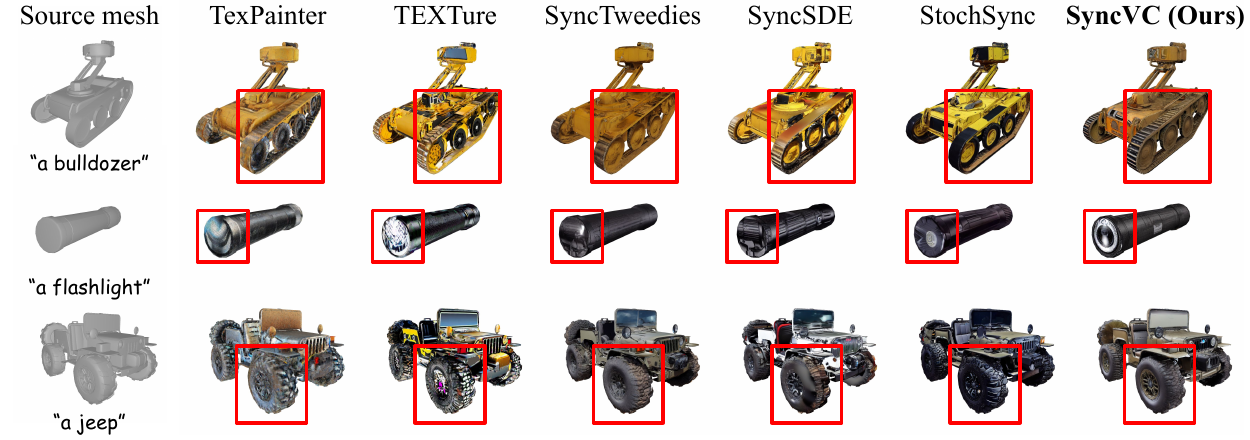}
	\vspace{-6mm}
	\caption{[Best viewed when magnified.]  \textbf{Our method outperforms baselines on 3D mesh texturing by producing artifact-free and realistic textures.} 
    We emphasize that \methodname~well preserves fine details such as the chain structure on the bulldozer tracks (Row 1), detailed view of the flashlight’s front lens (Row 2), and the overall natural appearance of the vehicle, including fine-grained textures on the tires (Row 3), while baselines generate over-smoothed and unrealistic textures. 
    }
\label{fig:mesh_cmp}
\end{figure}

Table~\ref{tab:mesh} shows that the proposed method outperforms all baselines. 
As illustrated in {Figure~\ref{fig:mesh_cmp}}, baselines often exhibit artifacts and inconsistent textures, whereas our method produces high-quality and detailed textures while alleviating such artifacts. 
These results demonstrate that \methodname~generalizes well across different modalities and dimensional settings, highlighting its effectiveness as a fundamental framework for collaborative generation.

\begin{table}[h!]
\caption{
Quantitative evaluation on text-guided 3D mesh texturing.
\methodname~shows superior performance across all baselines in terms of distributional alignment (FID~\cite{heusel2017gans}, KID~\cite{binkowski2018demystifying}), and comparable results on image-text alignment (CLIP-S~\cite{radford2021learning}).
KID score is scaled by $10^3$.
}
\vspace{-2mm}
\centering
\setlength{\tabcolsep}{4pt} 
\scalebox{0.8}{
    \begin{tabular}{l c c c c c c}
    \toprule
    Method & TexPainter~\cite{zhang2024texpainter} & TEXTure~\cite{richardson2023texture} &
    SyncTweedies~\cite{kim2024synctweedies} & SyncSDE~\cite{lee2025syncsde}  & StochSync~\cite{yeo2025stochsync} & 
    \textbf{\methodname~(Ours)}   \\
    \midrule
    FID~\cite{heusel2017gans} $\downarrow$ & 192.08 & 188.26 & 163.08 & 172.74 & \underline{162.71} & \textbf{161.63}  \\
    KID~\cite{binkowski2018demystifying} $\downarrow$ & 110.84 & 96.28 & 82.05 & 85.12 & \underline{80.11} & \textbf{75.17}  \\
    CLIP-S~\cite{radford2021learning} $\uparrow$ & 0.283 & 0.288 & \textbf{0.292} & 0.288 & 0.289 & \underline{0.290}  \\
    \bottomrule
    \end{tabular}
    }
    \vspace{-4mm}
    \label{tab:mesh}
\end{table}

\subsection{Additional analysis}

\paragraph{Effectiveness of \methodname~on extreme scenarios.}

\begin{figure}[h!]
	\centering
    	\includegraphics[width=1.0\linewidth]{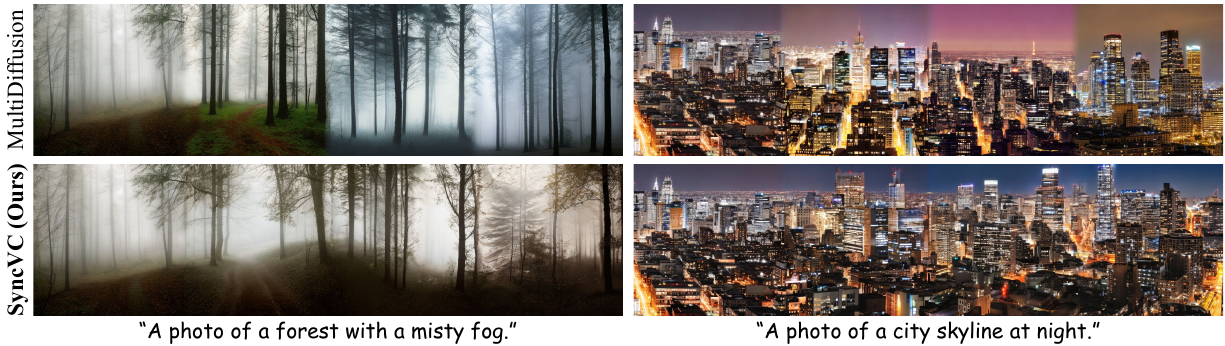}
	\vspace{-6mm}
	\caption{
    \textbf{Our method shows superior performance in wide image generation under an extreme small-overlap setting} (16 pixels, 3.125\% of patch width). 
    \methodname~maintains coherent style and consistent colors across patches, whereas MultiDiffusion~\cite{bar2023multidiffusion} fails to produce visually consistent results. 
    This result stems from introducing variational controls, which more effectively models complex correlations between trajectories than heuristic approximations used in baselines.
    }
\label{fig:wide_image_small_overlap}
\end{figure}

We further consider a more challenging wide image generation setting that uses a very small overlap of 16 pixels, to show the effectiveness of \methodname~under extreme conditions.
This setting is also practically important, as smaller overlaps require fewer trajectories for the same resolution, thereby reducing computational cost.
Under this setting, we compare our method with MultiDiffusion~\cite{bar2023multidiffusion}, which achieves the best performance among the baselines (see Table~\ref{tab:wide_image}).
As shown in Figure~\ref{fig:wide_image_small_overlap}, baselines exhibit significant performance degradation, whereas our method maintains strong style consistency.-
We attribute this robustness to the coupling kernel introduced in Eq.~\ref{eq:elbo_simplified}, since marginalizing the control variables yields a multimodal distribution $q(\rvx_{t-1}^{(n)}\mid \rvx_t^{(n)}, \rvx_t^{(n-1)}, \vy)$ that is capable of modeling complex relationships between trajectories.

\begin{figure}[h!]
    \vspace{-2mm}
	\centering
    	\includegraphics[width=1.0\linewidth]{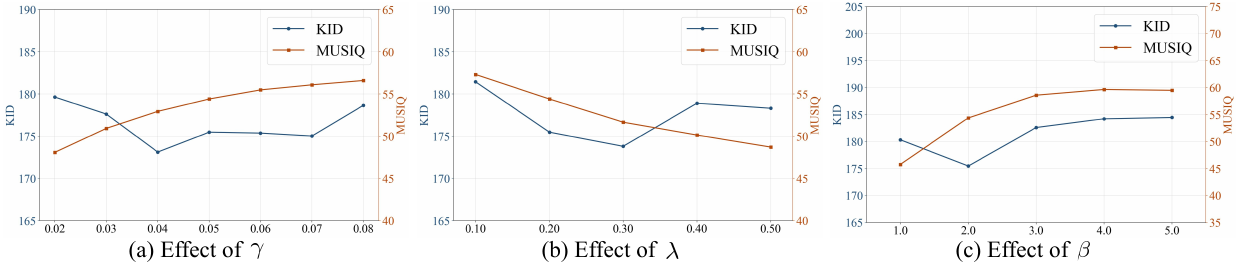}
	\vspace{-6mm}
	\caption{
    \textbf{Effects of hyperparameters ($\gamma$, $\lambda$, $\beta$) on the optical illusion generation task.}
    These values offer a trade-off between jointly capturing both semantics (KID) and visual quality (MUSIQ).
    }
    \vspace{-3mm}
\label{fig:analysis_misc}
\end{figure}

\paragraph{Effects of hyperparameters.}
Our parameterization contains three tunable coefficients: the weight of the reward function ($\gamma$), the weight of the KL-divergence term in ELBO ($\lambda$), and the control strength ($\beta$).
Figure~\ref{fig:analysis_misc} (a) shows the effects of these coefficients in optical illusion generation.
Firstly, as $\gamma$ increases from a small value, it better captures both semantics simultaneously, leading to improved KID scores.
However, excessively large $\gamma$ over-constrains the 2nd trajectory to resemble the 1st one, causing only one semantic to dominate. 
As a result, although the visual quality (MUSIQ) may improve, both semantics are no longer jointly captured and KID degrades.
Secondly, increasing $\lambda$ regularizes the 2nd trajectory toward the original diffusion prior, thereby reducing the influence of the 1st trajectory. 
This produces an effect analogous to decreasing $\gamma$. 
Lastly, increasing $\beta$ initially improves the overall quality as the controls strongly guide the trajectory to satisfy the objective in Eq.~\ref{eq:practical_objective_gaussian}.
However, too large $\beta$ may weaken the ability to jointly capture both semantics.

%


\section{Conclusion}
\label{sec:conclusion}

In this work, we propose a principled framework for collaborative generation based on optimal control. 
Unlike prior approaches that rely heavily on heuristic designs, our method is derived from a mathematically grounded formulation, providing a novel perspective on diffusion synchronization.
The proposed method demonstrates strong performance across diverse collaborative generation tasks, establishing a promising direction for extending pretrained generative priors to more versatile settings.
Despite these advantages, our approach has several limitations.
Because it relies on test-time optimization, it incurs additional computational cost compared to optimization-free approaches, motivating the development of more efficient guidance strategies. 
Furthermore, the formulation is currently restricted to differentiable rewards; 
extending it to incorporate non-differentiable objectives would enable broader applicability across diverse scenarios.

\begin{acksection}
We thank Justus Will and Jan Groeneveld for additional discussions and feedback. 
Stephan Mandt acknowledges funding from the National Science Foundation (NSF) through an NSF CAREER Award IIS-2047418, IIS2007719, the NSF LEAP Center.
\end{acksection}

{
    \small
    \bibliographystyle{plain}
    \bibliography{main}
}

\clearpage


\appendix

\section{Derivation of the ELBO (Eq.~\ref{eq:elbo_simplified})}
\label{sec:supp_method}

Let $\mU := \{\rvu_t^{(n-1)}\}_{n=2,\,t=1}^{N,\,T}$. The joint generative model factorizes as
\begin{equation}
    p(\rvx_{0:T}^{(1:N)}, \vy) = p(\vy \mid \rvx_0^{(1:N)}) \prod_{n=1}^{N} p(\rvx_T^{(n)}) \prod_{n=1}^{N} \prod_{t=1}^{T} p_\vphi(\rvx_{t-1}^{(n)} \mid \rvx_t^{(n)}),
    \label{eq:joint_app}
\end{equation}
with $\log p(\vy \mid \rvx_0^{(1:N)}) = r(\vy, \mX) - \log Z(\vy)$ from Eq.~\ref{eq:likelihood_def_via_energy}. Augmenting the variational distribution of Eq.~\ref{eq:variational_q} with the controls gives
\begin{equation}
    q(\rvx_{0:T}^{(1:N)} \mid \vy;\, \mU) = q(\rvx_{0:T}^{(1)}) \prod_{n=2}^{N} q(\rvx_T^{(n)}) \prod_{n=2}^{N} \prod_{t=1}^{T} q(\rvx_{t-1}^{(n)} \mid \rvx_t^{(n)}, \rvx_t^{(n-1)}, \rvu_t^{(n-1)}, \vy).
    \label{eq:variational_aug}
\end{equation}
The first trajectory is sampled from the pretrained prior and reverse sampling is initialized at the prior's terminal Gaussian; we therefore set $q(\rvx_{0:T}^{(1)}) = p(\rvx_{0:T}^{(1)})$ and $q(\rvx_T^{(n)}) = p(\rvx_T^{(n)})$ for $n \geq 2$.
Jensen's inequality yields
\begin{equation}
    \log p(\vy) \geq \EE_q\!\left[\log p(\rvx_{0:T}^{(1:N)}, \vy) - \log q(\rvx_{0:T}^{(1:N)} \mid \vy;\, \mU)\right].
    \label{eq:jensen_app}
\end{equation}
Substituting Eqs.~\ref{eq:joint_app} and~\ref{eq:variational_aug} and cancelling the factors that coincide under the assumed $q$,
\begin{align}
    \log p(\rvx_{0:T}^{(1:N)}, \vy) - \log q(\rvx_{0:T}^{(1:N)} \mid \vy;\, \mU)
    &= \log p(\vy \mid \rvx_0^{(1:N)}) \nonumber \\
    &\quad + \sum_{n=2}^{N} \sum_{t=1}^{T} \log \frac{p_\vphi(\rvx_{t-1}^{(n)} \mid \rvx_t^{(n)})}{q(\rvx_{t-1}^{(n)} \mid \rvx_t^{(n)}, \rvx_t^{(n-1)}, \rvu_t^{(n-1)}, \vy)}.
    \label{eq:logratio_app}
\end{align}
The likelihood term contributes $\EE_q[r(\vy, \mX)] - \log Z(\vy)$, while the tower property turns each transition log-ratio into a negative KL divergence under the marginal $q(\rvx_t^{(n)}, \rvx_t^{(n-1)})$. Combining with Eq.~\ref{eq:jensen_app},
\begin{align}
    \log p(\vy)
    &\geq \EE_q[r(\vy, \mX)] - \log Z(\vy) \nonumber \\
    &\quad - \sum_{n=2}^{N} \sum_{t=1}^{T} \KL\!\left(q(\rvx_{t-1}^{(n)} \mid \rvx_t^{(n)}, \rvx_t^{(n-1)}, \rvu_t^{(n-1)}, \vy) \,\|\, p_\vphi(\rvx_{t-1}^{(n)} \mid \rvx_t^{(n)})\right).
    \label{eq:bound_app}
\end{align}
Since $\log Z(\vy)$ is independent of $\mU$, optimizing this bound is equivalent to optimizing the objective in Eq.~\ref{eq:elbo_simplified}, with $\lambda$ generalizing the unit weighting on the KL terms. \qed

\section{Derivation of the objective for DDIM (Eq.~\ref{eq:practical_objective_ddim})}
\label{sec:supp_ddim_derivation}

Under the DDIM~\cite{song2020denoising} parameterization, $\vmu_\vphi\!\left(\rvx_t^{(n)}, t\right)$ is defined as follows:
\begin{equation}
    \vmu_\vphi\!\left(\rvx_t^{(n)}, t\right) = \underbrace{\sqrt{\frac{\alpha_{t-1}}{\alpha_t}}}_{a_t} \rvx_t^{(n)} + \underbrace{\left( \sqrt{1-\alpha_{t-1} }  - \sqrt{\frac{(1-\alpha_t)\alpha_{t-1}}{\alpha_t}} \right)}_{b_t} \epsilon_{\theta}(\rvx_t^{(n)}, t).
    \label{eq:ddim_reverse}
\end{equation}
For $\bar{\vmu}_t^{(n)}$ in Eq.~\ref{eq:q_x_param}, we use an analogous parameterization:
\begin{equation}
    \bar{\vmu}_t^{(n)} = \sqrt{\frac{\alpha_{t-1}}{\alpha_t}} \bar{\rvx}_t^{(n)} + \left( \sqrt{1-\alpha_{t-1} }  - \sqrt{\frac{(1-\alpha_t)\alpha_{t-1}}{\alpha_t}} \right) \bar{\epsilon}_{\theta}(\bar{\rvx}_t^{(n)}, t),
\end{equation}
where $\bar{\epsilon}_{\theta}(\bar{\rvx}_t^{(n)}, t)$ is calculated using conditional score-based sampling~\cite{dhariwal2021diffusion, song2020score, lee2023conditional} as:
\begin{equation}
    \bar{\epsilon}_{\theta}(\bar{\rvx}_t^{(n)}, t) = \epsilon_{\theta}(\bar{\rvx}_t^{(n)}, t) + \frac{\gamma}{2} \sqrt{1-\alpha_t}
    \nabla_{\rvx_t^{(n)}} 
    \left\| f \left( \rvx_t^{(n-1)}, \vy  \right)
    - \bar{\rvx}_t^{(n)}
    \right\|_2^2.
\end{equation}
Using these parameterizations, we obtain the practical objective optimized with the DDIM sampler:
\begin{align}
    \mathcal{J} & =  - \tilde{r}\!\left(\vy, \hat{\rvx}_{0|t}^{(1)}, \dots, \hat{\rvx}_{0|t}^{(n)}\right)
    + \lambda \left\| \bar{\vmu}_t^{(n)} - \vmu_\vphi\!\left(\rvx_t^{(n)}, t\right) \right\|_2^2 \nonumber \\
    & \leq  - \tilde{r}\!\left(\vy, \hat{\rvx}_{0|t}^{(1)}, \dots, \hat{\rvx}_{0|t}^{(n)}\right)
    + \lambda \left\| a_t \beta \rvu_t^{(n-1)}  + b_t (\bar{\epsilon}_{\theta}(\bar{\rvx}_t^{(n)}, t) - {\epsilon}_{\theta}(\bar{\rvx}_t^{(n)}, t) )\right\|_2^2 \nonumber \\
    & \leq  - \tilde{r}\!\left(\vy, \hat{\rvx}_{0|t}^{(1)}, \dots, \hat{\rvx}_{0|t}^{(n)}\right)
    + \lambda a_t^2 \beta^2 \left\| \rvu_t^{(n-1)} \right\|_2^2  \nonumber \\
    &\quad + \lambda b_t^2 \left\|
    \epsilon_{\theta}(\bar{\rvx}_t^{(n)} , t)
    + \frac{\gamma}{2} \sqrt{1-\alpha_t}
    \nabla_{\rvx_t^{(n)}} 
    \left\| f \left( \rvx_t^{(n-1)}, \vy  \right)
    - \bar{\rvx}_t^{(n)}
    \right\|_2^2
    - \epsilon_{\theta}(\rvx_t^{(n)}, t)
    \right\|^2_2.
    \label{eq:practical_objective_ddim_full_derivation}
\end{align}
Here, we empirically omit the coefficient $\beta^2$ in the second term of Eq.~\ref{eq:practical_objective_ddim_full_derivation}, which improves performance and yields the final objective presented in Eq.~\ref{eq:practical_objective_ddim}.

\section{Pseudocode of \methodname}
\label{sec:supp_pseudocode}
Algorithm~\ref{alg:overview} provides a high-level overview of the practical implementation.
Here, we note that controls are optimized sequentially in a greedy manner.
Specifically, for each diffusion timestep $t$, instead of jointly optimizing $N-1$ control variables $\{ \rvu_t^{(1)}, \dots, \rvu_t^{(N-1)}\}$, we iterate over $n$ and optimize each control variable $\rvu_t^{(n)}$ using the previously optimized variables $\{ \rvu_t^{(1)}, \cdots, \rvu_t^{(n-1)} \}$.
The optimization objective in Eq.~\ref{eq:practical_objective_gaussian} is factorized to $N-1$ terms (\textit{i.e.}, each term in the summation) and $(n-1)$-th term is used as the optimization objective for $\rvu_t^{(n)}$.
We adopt this setting since we empirically observe that greedy optimization yields better results.


\begin{algorithm}[h!]
	\caption{\methodname}
	\label{alg:overview}
	\begin{algorithmic}[1]
		
		\State \textbf{Inputs:} Observation $\vy$, Noisy latent variables $\rvx_T^{(1)}, \dots, \rvx_T^{(N)}$, Denoising kernel $p_\vphi$, Optimization step $K$, Hyperparameters $\gamma$, $\lambda$, $\beta$

        \For{$t \gets T, \cdots , 1$}

            \State Initialize $\{ \rvu_t^{(1)}, \dots, \rvu_t^{(N-1)}\}$ as zero

            \State Calculate $\rvx_{t-1}^{(1)}$ using the denoising kernel $p_\vphi$
            
            \For{$n \gets 2, \cdots , N$}

                \For{$i \gets 1, \cdots, K$}
                
                    \State Calculate the $(n-1)$-th term of the objective in Eq.~\ref{eq:practical_objective_gaussian}   
    
                    \State Optimize $\rvu_t^{(n-1)}$ using the calculated optimization objective 
                
                \EndFor
                
                \State Calculate $\rvx_{t-1}^{(n)}$ using Eq.~\ref{eq:q_x_param} with optimized $\rvu_t^{(n-1)}$ and $\rvx_{t}^{(n)}$
    
            \EndFor
            
        \EndFor

        \State \textbf{Outputs:} Sequence $\mX = \{\rvx_0^{(1)}, \ldots,  \rvx_0^{(N)} \}$
        
	\end{algorithmic}
\end{algorithm}

\section{Additional results}
\label{sec:supp_exp_details}

\subsection{Wide image generation}

\paragraph{Experimental details.}

We generate $2048 \times 512$ sized wide images for evaluation, with each patch of size $512^2$.
For our method, five distinct trajectories are sampled with an overlap of 128 pixels.
The generated patches are then sequentially concatenated so that the patch from the $n$-th trajectory is placed on top of that from the $(n+1)$-th trajectory, yielding a single wide image.
Each control variable $\mathbf{u}_t^{(n)}$ is optimized for five steps with a learning rate of $10^{-2}$ using the Adam optimizer~\cite{kingma2014adam}.
Hyperparameters are set to $\beta=1.0$, $\gamma=2.5$, $\lambda=2.0$.
Across all methods, we use the pretrained Stable Diffusion v2.1-base~\cite{rombach2022high}\footnote{Accessed via https://huggingface.co/Manojb/stable-diffusion-2-1-base. CreativeML Open RAIL++-M License} with 50 steps of DDIM~\cite{song2020denoising} sampling and classifier-free guidance~\cite{ho2021classifierfree} to ensure a fair comparison.
For baselines, we run their official codes\footnote{MultiDiffusion: https://github.com/omerbt/MultiDiffusion} \footnote{SyncTweedies: https://github.com/KAIST-Visual-AI-Group/SyncTweedies, MIT License} \footnote{SyncSDE: https://github.com/hjl1013/SyncSDE} \footnote{StochSync: https://github.com/KAIST-Visual-AI-Group/StochSync, MIT License}.
In addition, we follow the baseline setup for noise initialization.
Instead of independently sampling noise for each patch, a single wide latent noise map is first sampled from a Gaussian distribution, then cropped into the corresponding patch regions for each trajectory.
We adopt the same process to ensure fairness in comparison.

To measure Intra-LPIPS~\cite{zhang2018unreasonable}, Intra-Style-Loss~\cite{gatys2016image}, $\chi^2$-Histogram distance, and Histogram intersection, each wide image is cropped into four non-overlapping views of size $515^2$, and the distances over all pairwise combinations (which is 6) are calculated.
Note that the color histograms are computed in the HSV space.
KID~\cite{binkowski2018demystifying} score is calculated using randomly cropped $512^2$ views from each wide image.
These five metrics are measured across all prompts and then averaged.
Reference images for KID measurement are constructed by generating 1,000 images per prompt using the pretrained Stable Diffusion v1.5~\cite{rombach2022high}\footnote{Accessed via https://huggingface.co/stable-diffusion-v1-5/stable-diffusion-v1-5. CreativeML OpenRAIL M License}.

\paragraph{Alternative reward functions.}
We further demonstrate that the proposed method can accommodate different reward designs by considering a variant with an additional semantic guidance term. 
Specifically, we augment the reward in Eq.~\ref{eq:wide_image_reward} with the CLIP similarity~\cite{radford2021learning} between the generated patch $\rvx_0^{(n)}$ and the text prompt $\vy$. 
The resulting reward is defined as
\begin{equation}
    \tilde{r}\!\left(\vy, \rvx_0^{(1)}, \dots, \rvx_0^{(n)}\right) = - \frac{\gamma}{2} \left\| \mathbf{M} \odot \left( f\!\left(\rvx_0^{(n-1)}\right) - \rvx_0^{(n)} \right) \right\|^2_2 + \lambda^{\mathrm{clip}} \cdot \mathrm{Sim} \left(f^{\mathrm{img}}(\rvx_0^{(n)}), f^{\mathrm{txt}}(\vy) \right),
    \label{eq:wide_image_reward_clip}
\end{equation}
where $f^{\mathrm{img}}$ and $f^{\mathrm{txt}}$ denote the image and text encoders of the pretrained CLIP model, respectively, $\mathrm{Sim}(\cdot, \cdot)$ denotes cosine similarity, and $\lambda^{\mathrm{clip}}$ is a scalar hyperparameter.
We select $\lambda^{\mathrm{clip}}=0.05$.

We apply the CLIP-based guidance term to all patches, \textit{i.e.}, for $1 \leq n \leq N$. 
Since the first patch $\rvx_t^{(1)}$ is also guided by the CLIP-based term, we introduce an additional control variable $\rvu_t^{(0)}$ for this patch. 
Consequently, the ELBO includes an additional KL term associated with $\rvu_t^{(0)}$, yielding 
\begin{align}
    \gL(\vy) :=\; \EE_{q} [r\!\left(\vy, \mX\right)]
    &- \lambda \sum_{n=2}^{N} \sum_{t=1}^{T} \! \KL\!\left( q\!\left(\rvx_{t-1}^{(n)} \mid \rvx_t^{(n)}, \rvx_t^{(n-1)}, \rvu_t^{(n-1)}, \vy\right) \,\|\, p\!\left(\rvx_{t-1}^{(n)} \mid \rvx_t^{(n)}\right) \right) \nonumber \\ 
    &- \lambda \sum_{t=1}^{T} \! \KL\!\left( q\!\left(\rvx_{t-1}^{(1)} \mid \rvx_t^{(1)}, \rvu_t^{(0)}, \vy\right) \,\|\, p\!\left(\rvx_{t-1}^{(1)} \mid \rvx_t^{(1)}\right) \right) .
    \label{eq:elbo_simplified_additional_condition}
\end{align}
We report the quantitative results using an alternative reward function from Eq.~\ref{eq:wide_image_reward_clip} in Table~\ref{tab:wide_image_clip} (See ``SyncVC*'' column).
As shown, this variant still outperforms the best baseline, MultiDiffusion~\cite{bar2023multidiffusion}, demonstrating that  our framework can accommodate different reward choices while maintaining strong performance.

\begin{table}[h!]
\caption{
Quantitative evaluation on wide image generation with an alternative reward choice.
SyncVC* denotes our method using the reward function in Eq.~\ref{eq:wide_image_reward_clip}. 
Our framework consistently outperforms the best baseline, MultiDiffusion~\cite{bar2023multidiffusion}, across all metrics, demonstrating its flexibility in incorporating different reward designs.
KID~\cite{binkowski2018demystifying} score is scaled by $10^3$.
}
\vspace{-2mm}
\centering
\setlength{\tabcolsep}{6.0pt} 
\scalebox{0.85}{
    \begin{tabular}{l c c c c c}
    \toprule
    Method & MultiDiffusion~\cite{bar2023multidiffusion}  &  \textbf{\methodname~(Ours, Eq.~\ref{eq:wide_image_reward})} & \textbf{\methodname*~(Ours, Eq.~\ref{eq:wide_image_reward_clip})}   \\
    \midrule
    Intra-LPIPS~\cite{zhang2018unreasonable} $\downarrow$ & 0.637 & \textbf{0.592} & \underline{0.594} \\
    Intra-Style-Loss~\cite{gatys2016image} $\downarrow$  &  {58.46} & \textbf{44.34} & \underline{47.13}  \\
    $\chi^2$-Histogram dist. $\downarrow$ &  {1.211}  & \textbf{0.751}  & \underline{0.784} \\
    Histogram intersect. $\uparrow$ & {0.549}  & \textbf{0.665}   & \underline{0.657} \\
    KID~\cite{binkowski2018demystifying} $\downarrow$ & 58.26 & \textbf{52.07}  & \underline{53.43}  \\
    \bottomrule
    \end{tabular}
    }
    \label{tab:wide_image_clip}
\end{table} 

\clearpage

\paragraph{Details on style guidance.}
We incorporate style guidance by adding a style transfer loss between a style reference image and $\mathbf{x}_0^{(n)}$ $(1 \leq n \leq N)$ within the reward function of Eq.~\ref{eq:wide_image_reward}.
Since the first patch $\rvx_t^{(1)}$ is also subject to the style guidance, the ELBO for stylized wide-image generation is identically derived as Eq.~\ref{eq:elbo_simplified_additional_condition}.

Style references used for experiments are borrowed from the internet\footnote{https://github.com/gordicaleksa/pytorch-neural-style-transfer, MIT license}.
Specifically, following Gatys et al.~\cite{gatys2016image}, the style transfer loss is defined as a weighted sum of a content loss and a style loss.
Let $\mathbf{F} (\mathbf{I}) \in \mathbb{R}^{C\times M}$ denote the features of an image $\mathbf{I}$ extracted using the pretrained VGG network~\cite{simonyan2014very}, where $C$ is the total number of feature maps and $M$ is the spatial resolution of each feature map.
The content loss is defined as the squared error between features of two images:
\begin{equation}
    \mathcal{L}_{\text{content}}(\mathbf{I}_1, \mathbf{I}_2) = \frac{1}{2} \|\mathbf{F}(\mathbf{I}_1) - \mathbf{F} (\mathbf{I}_2) \|_F^2
\end{equation}
where $\mathbf{I}_1$ denotes the content image and $\mathbf{I}_2$ is the stylized image. 
In the stylized wide image generation scenario, the content image is defined as $\mathbf{x}_0^{(n)}$ obtained without optimizing controls.
For the style representation of image $\mathbf{I}$, we use the Gram matrix $\mathbf{G} (\mathbf{I}) \in \mathbb{R}^{C \times C}$:
\begin{equation}
    \mathbf{G} (\mathbf{I})[i, j] = \sum_k \mathrm{Vec}(\mathbf{F}_{ik}(\mathbf{I}) ) \mathrm{Vec}( \mathbf{F}_{jk}  (\mathbf{I})) \quad \text{for} \quad 1 \leq i, j \leq C,
\end{equation}
where $\mathrm{Vec}(\cdot)$ stands for the vectorization operation of a given matrix.
The style loss is then defined as
\begin{equation}
    \mathcal{L}_{\text{style}}(\mathbf{I}_2, \mathbf{I}_3) = \frac{1}{4 C^2 M^2} \sum_{i,j} \left( G_{ij} (\mathbf{I}_2) - G_{ij} (\mathbf{I}_3) \right)^2,
\end{equation}
where $\mathbf{I}_3$ denotes the style reference image.
In practice, we use features from multiple layers of VGG network to calculate style loss and use the averaged value.
The style transfer loss is finally given by
\begin{equation}
    \mathcal{L}_{\text{style-transfer}} = w_{\mathrm{content}} \mathcal{L}_{\text{content}} + w_{\mathrm{style}} \mathcal{L}_{\text{style}}.
\end{equation}
We choose $w_{\mathrm{content}} = 1.0$ and $w_{\mathrm{style}} = 10^{-4}$.

\paragraph{Results on extreme scenarios.}

\begin{figure}[h!]
	\centering
    	\includegraphics[width=1.0\linewidth]{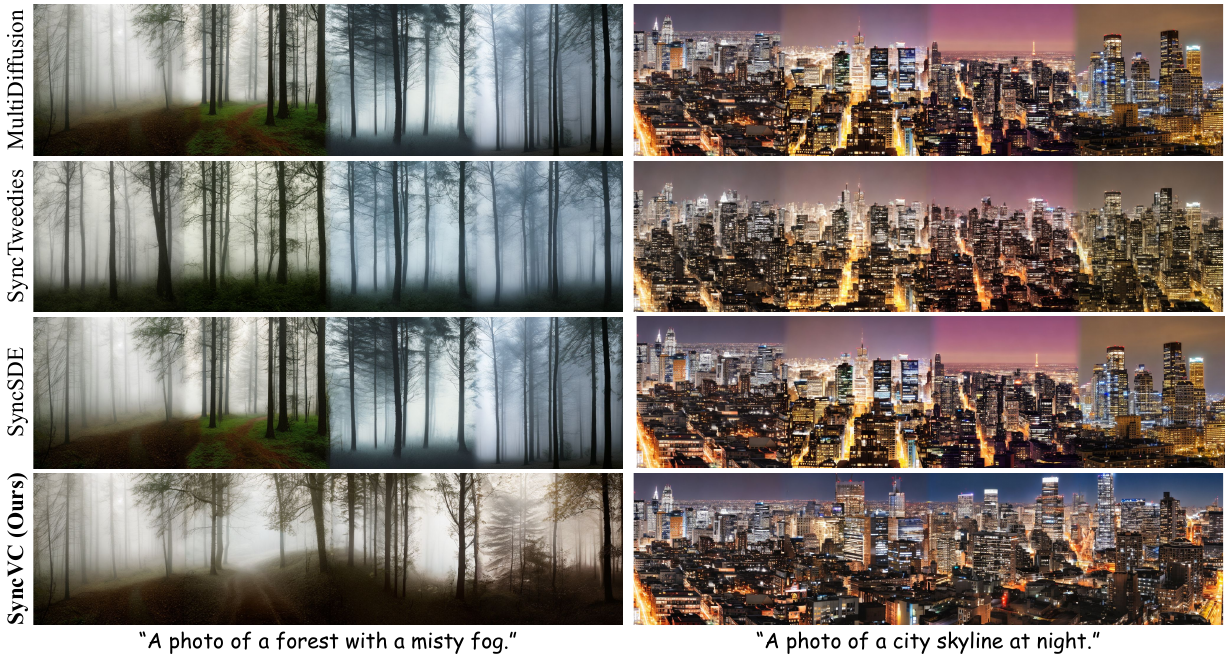}
	\vspace{-6mm}
	\caption{
    \textbf{Our method demonstrates superior performance in wide image generation under a small-overlap setting}, maintaining strong style and color consistency across the horizontal axis. 
    All baseline methods exhibit significant color changes.  
    }
\label{fig:supp_wide_image_small_overlap}
\end{figure}

For the sample shown in Figure~\ref{fig:wide_image_small_overlap} of the main paper, we additionally visualize the full figure including the results of SyncTweedies~\cite{kim2024synctweedies} and SyncSDE~\cite{lee2025syncsde} in Figure~\ref{fig:supp_wide_image_small_overlap}.
This corresponds to the wide image generation setting with a small overlap of only 16 pixels (over the patch width of 512 pixels).
As shown, baselines exhibit noticeable color inconsistencies between patches, while the proposed method produces consistent outputs.

\paragraph{Additional qualitative results.}

\begin{figure}[h!]
	\centering
    	\includegraphics[width=1.0\linewidth]{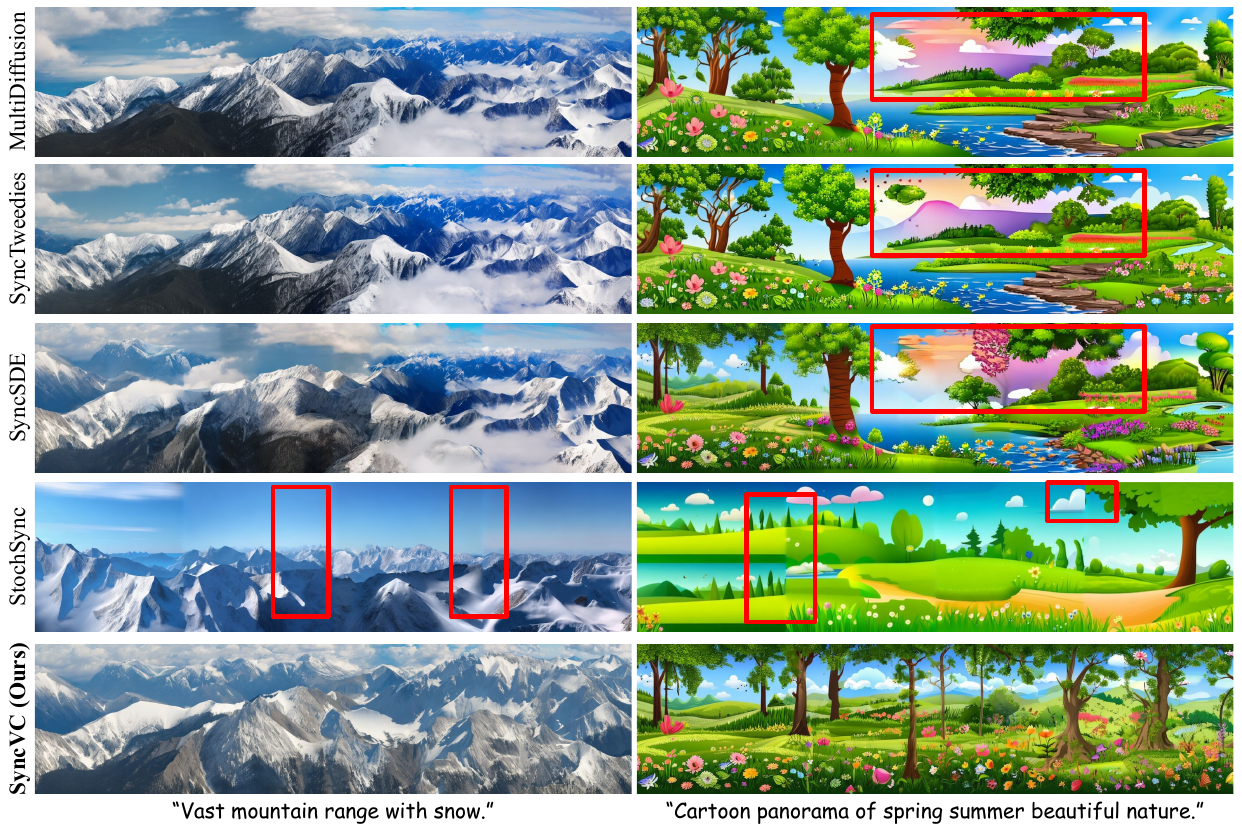}
	\vspace{-6mm}
	\caption{
    \textbf{Our method shows superior performance in wide image generation.} 
    (Left) \methodname~maintains a unified color and style, while baselines suffer from varying mountain and sky colors, or discontinuities (see bounding box).
    (Right) Our method generates cartoon-like panorama with consistent styles of tree and flowers, while baselines result in artifacts with inconsistent colors or discontinuities (see bounding box). 
    }
\label{fig:supp_wide_image_cmp}
\end{figure}

\begin{figure}[t!]
	\centering
    	\includegraphics[width=1.0\linewidth]{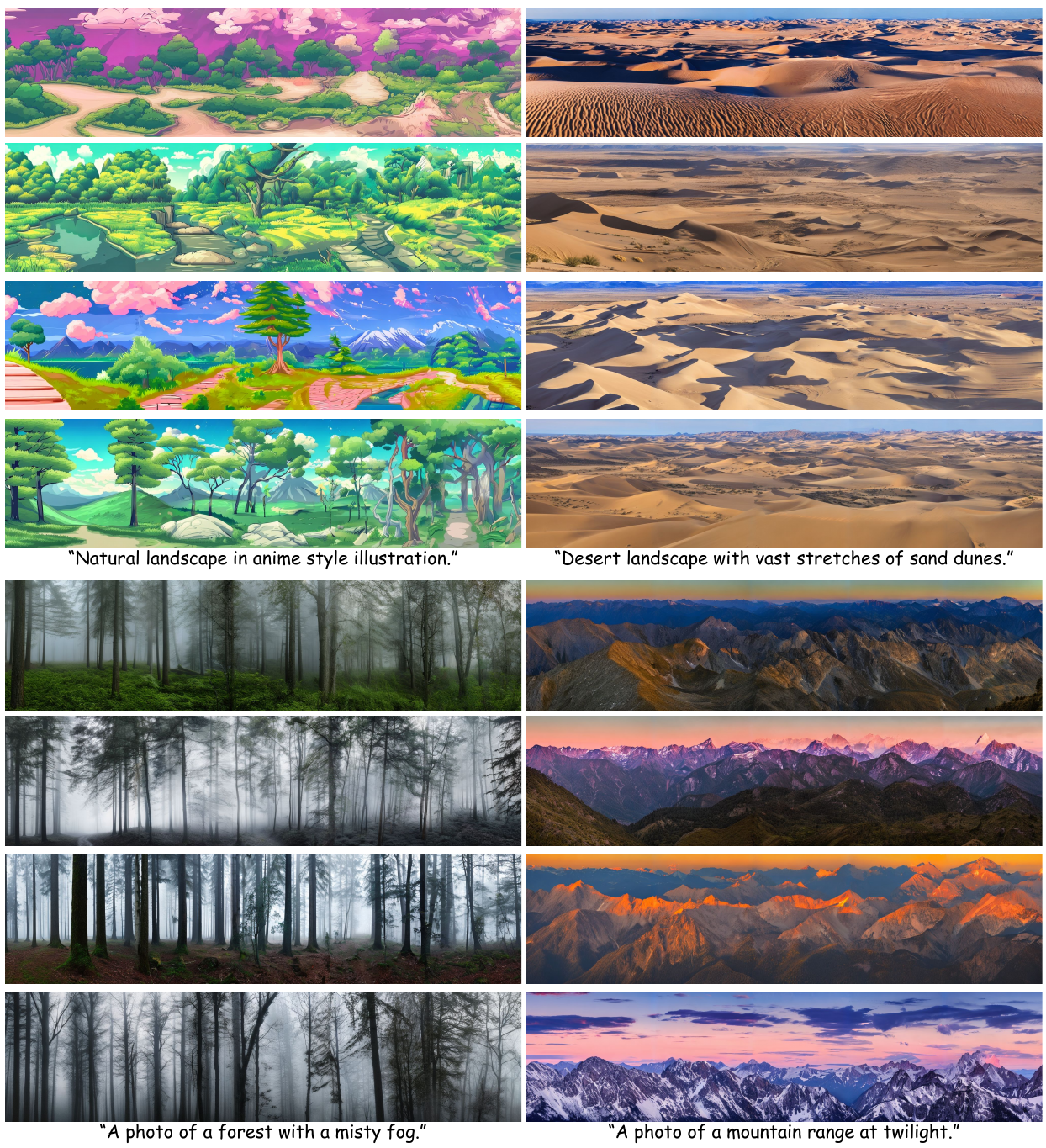}
	\vspace{-6mm}
	\caption{
    \textbf{Our method generates high-quality wide images conditioned on diverse text prompts.} 
    We present multiple wide image samples generated using the pretrained Stable Diffusion~\cite{rombach2022high} for various text prompts, all exhibiting strong style consistency.
    }
\label{fig:supp_wide_image_ours}
\end{figure}

\begin{figure}[t!]
	\centering
    	\includegraphics[width=1.0\linewidth]{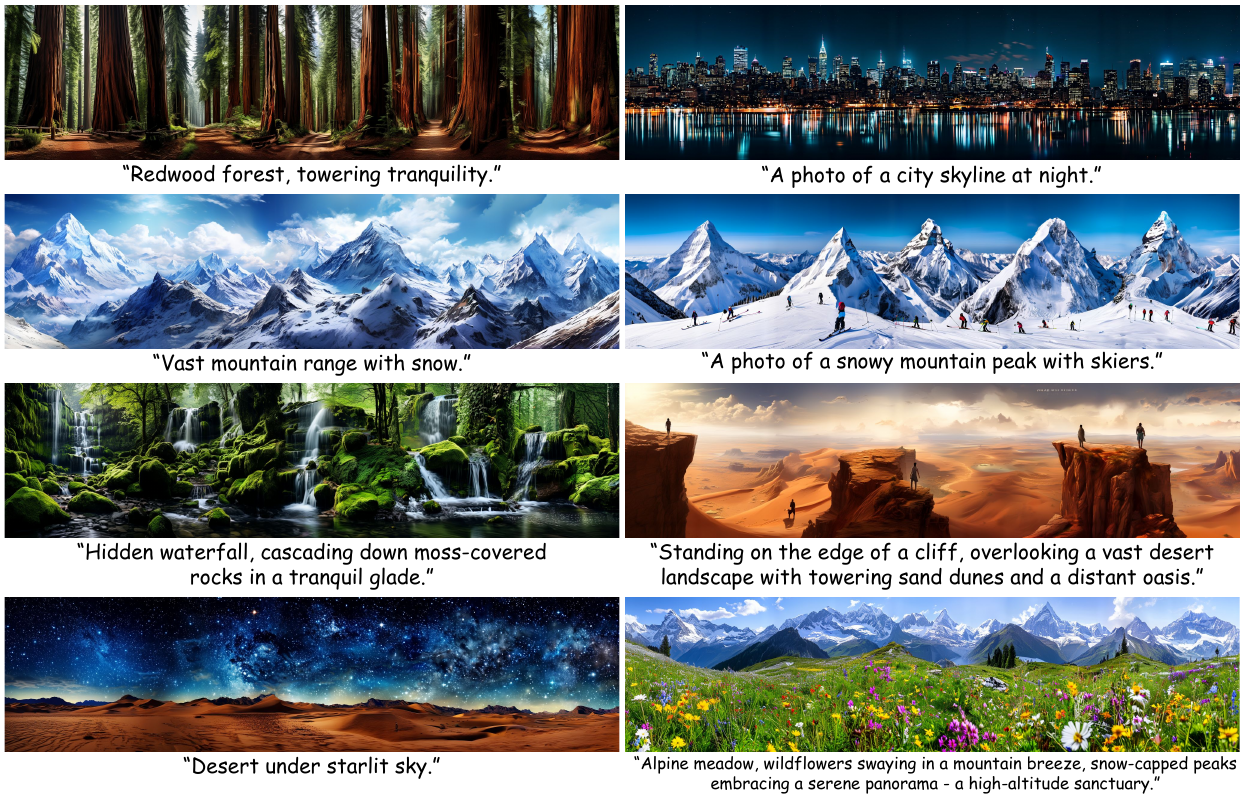}
	\vspace{-6mm}
	\caption{
    \textbf{Our method can also synthesize high-resolution wide image when combined with SANA model~\cite{xie2025sana}.} 
    We visualize various wide images at the resolution of $4096 \times 1024$ using the pretrained SANA model, where each generated patch has a resolution of $1024^2$.
    }
\label{fig:supp_wide_image_ours_sana_1}
\end{figure}

\begin{figure}[t!]
	\centering
    	\includegraphics[width=1.0\linewidth]{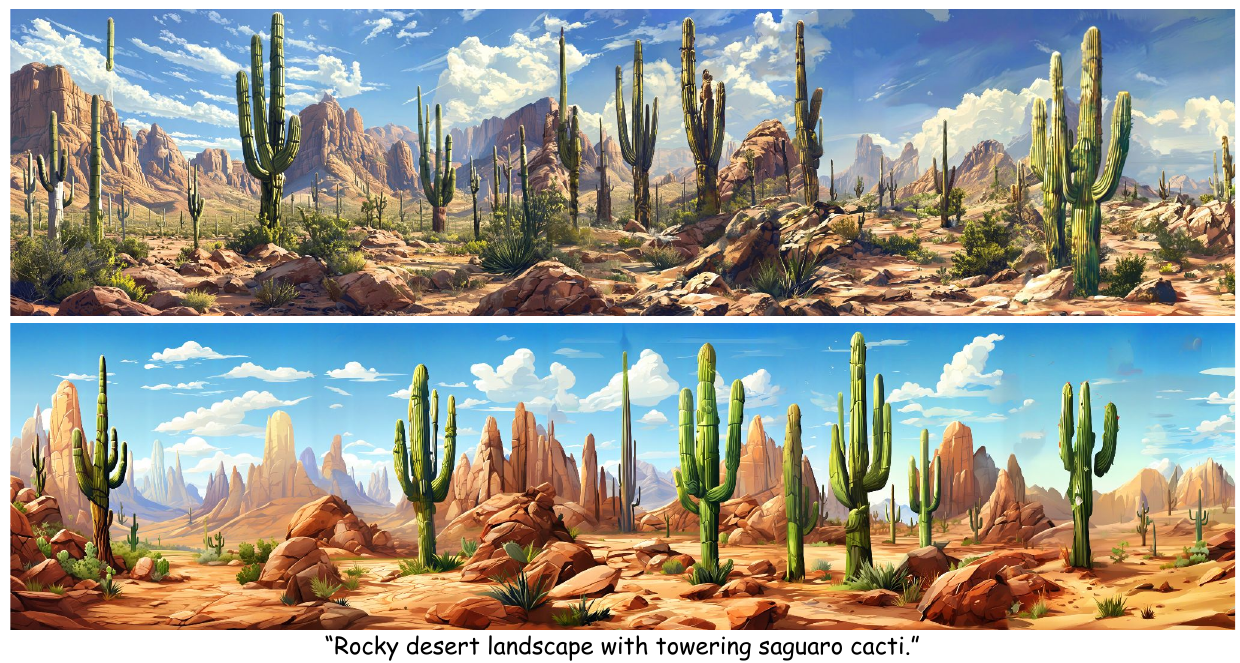}
	\vspace{-6mm}
	\caption{
    \textbf{Our method is even capable of generating $8192 \times 2048$-sized wide image.}
    We use the pretrained SANA model~\cite{xie2025sana} that generates patches at a resolution of $2048^2$, and extend it along the horizontal axis using our method to generate ultra-high-resolution images.
    }
\label{fig:supp_wide_image_ours_sana_2}
\end{figure}

We show additional qualitative comparisons with baselines in Figure~\ref{fig:supp_wide_image_cmp}, and additional results of our method in Figure~\ref{fig:supp_wide_image_ours}.
Furthermore, we show that the proposed method can be also applied to recent generative model with stronger priors that synthesize high-resolution images. 
Specifically, we use the pretrained SANA model~\cite{xie2025sana}\footnote{Accessed via https://huggingface.co/Efficient-Large-Model/Sana\_1600M\_1024px\_diffusers, NVIDIA License} \footnote{Accessed via https://huggingface.co/Efficient-Large-Model/Sana\_1600M\_2Kpx\_BF16\_diffusers, NVIDIA License} to generate wide images with the resolution of $4096\times1024$ and $8192 \times 2048$, and visualize it in Figure~\ref{fig:supp_wide_image_ours_sana_1} and~\ref{fig:supp_wide_image_ours_sana_2}.

\clearpage


\clearpage

\subsection{Optical illusion generation}

\paragraph{Experimental details.}

We generate images of $256^2$ resolution for all methods.
Each image is associated with two trajectories, corresponding to two different (transformation, prompt) pairs.
Specifically, we use the pretrained two-stage DeepFloyd-IF model~\cite{DeepFloydIF}, where the first and second stage models are IF-I-M-v1.0\footnote{Accessed via https://huggingface.co/DeepFloyd/IF-I-M-v1.0, DeepFloyd IF License Agreement} and IF-II-M-v1.0\footnote{Accessed via https://huggingface.co/DeepFloyd/IF-II-M-v1.0, DeepFloyd IF License Agreement}, respectively.
For baselines, we run their official implementation\footnote{Anagram-MTL: https://github.com/Pixtella/Anagram-MTL, Apache-2.0 License} for experiments.
For our method, guidance is applied only during the first-stage sampling. 
We use 30 steps of DDIM~\cite{song2020denoising} reverse process with classifier-free guidance~\cite{ho2021classifierfree}, and the noise scale parameter for second stage of DeepFloyd-IF model is fixed to 50 across all methods.
Each control variable $\mathbf{u}_t^{(n)}$ is optimized for five iterations with a learning rate of $10^{-2}$ using the Adam optimizer~\cite{kingma2014adam}. 
We use $\beta = 2.0$, $\gamma = 0.05$, and $\lambda = 0.2$ as default hyperparameters.

For evaluation, two views are sampled from each generated image. 
The 2nd view is obtained directly from the second trajectory, while the 1st view is constructed by applying the inverse illusion transformation to the second view (\textit{e.g.}, a counterclockwise rotation for a clockwise illusion transformation). 
Note that each view is associated with its corresponding prompt.
FID~\cite{heusel2017gans} and KID~\cite{binkowski2018demystifying} values are computed between the images of each view and the reference images for each (transformation, prompt) pair, then averaged.
We generate the reference images by synthesizing 1,000 images per prompt using the pretrained Stable Diffusion v1.5~\cite{rombach2022high}.
Furthermore, MUSIQ~\cite{ke2021musiq} is computed for both views, and the scores are averaged over all generated images.

\paragraph{Additional qualitative results.}

\begin{figure}[h!]
	\centering
    	\includegraphics[width=1.0\linewidth]{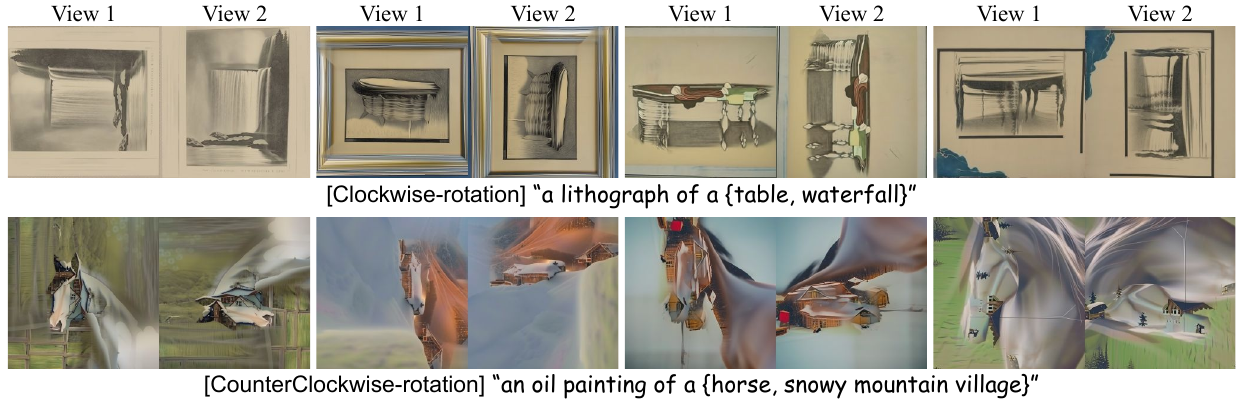}
	\vspace{-6mm}
	\caption{
    \textbf{Our method shows superior performance on the optical illusion generation task} by clearly incorporating two semantics specified by different text prompts.
    (Row 1) The generated image can be viewed as both a table and a waterfall under clockwise rotation.
    (Row 2) Each view encodes both a horse and a snowy mountain village under counterclockwise rotation.
    }
    \vspace{-3mm}
\label{fig:supp_illusion_ours}
\end{figure}

In Figure~\ref{fig:supp_illusion_ours}, we show the results of our method on two additional illusion transformations.
\methodname~generates high-resolution images that successfully encode both semantics under the illusion transformation.

\begin{figure}[h!]
	\centering
    	\includegraphics[width=1.0\linewidth]{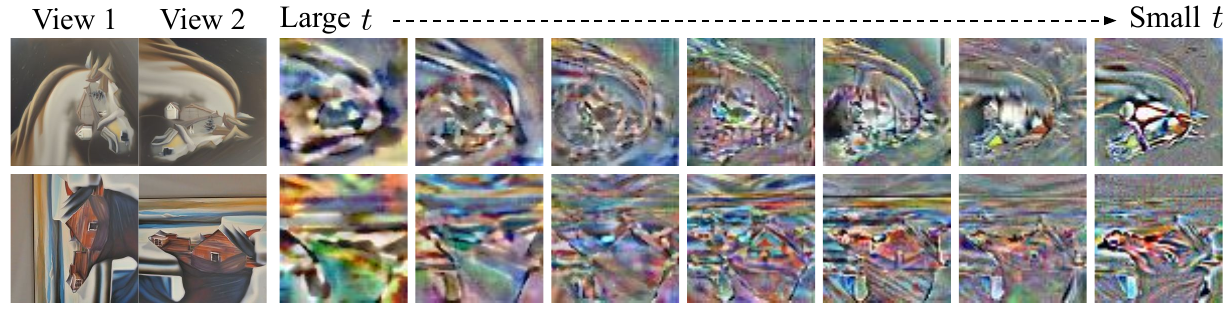}
	\caption{
    \textbf{Visualization of optimized controls.}
    The controls first capture coarse and low-level structures, then refine high-level features.
    We use the text prompts of ``an oil painting of a horse'' and ``an oil painting of a snowy mountain village'', with clockwise rotation.
    }
    \vspace{-4mm}
\label{fig:supp_control_visualizaiton}
\end{figure}

\paragraph{Visualization of controls.}
To provide intuition on the role of controls, we visualize the optimized control variables in Figure~\ref{fig:supp_control_visualizaiton}.
At early timesteps (large $t$), the controls focus on shaping the overall semantics of an image to satisfy the optimization objective, whereas at later timesteps (small $t$), they progressively manipulate fine-grained details.

\subsection{Text-guided 3D mesh texturing}

\paragraph{Experimental details.}

We use the pretrained Stable Diffusion v1.5~\cite{rombach2022high} with the pretrained depth-conditioned ControlNet~\cite{zhang2023adding}\footnote{Accessed via https://huggingface.co/lllyasviel/control\_v11f1p\_sd15\_depth, The CreativeML OpenRAIL M License} for synchronization-based methods~\cite{kim2024synctweedies, lee2025syncsde, yeo2025stochsync} (including ours), and Stable Diffusion v2-depth model\footnote{Accessed via https://huggingface.co/sd2-community/stable-diffusion-2-depth, CreativeML Open RAIL++-M License} for task-specific methods~\cite{zhang2024texpainter, richardson2023texture} following their original configuration.
Regarding the viewpoint setting, we fix the elevation to $15^\circ$ and uniformly sample eight azimuth angles from $[0^\circ, 360^\circ)$, resulting in eight diffusion trajectories. 
We use 8 DDIM~\cite{song2020denoising} steps, with the resolution of $768^2$ for each patch. 
Meanwhile, we follow the default viewpoint sampling and diffusion sampling configurations for baselines and run their official codes\footnote{TexPainter: https://github.com/Quantuman134/TexPainter, MIT License} \footnote{TEXTure: https://github.com/TEXTurePaper/TEXTurePaper, MIT License}.
For all methods, we prepend the prompt with the phrase ``Best quality, extremely detailed'' and use classifier-free guidance~\cite{ho2021classifierfree} with the negative prompt ``oversmoothed, blurry, depth of field, out of focus, low quality, bloom, glowing effect''.
To synthesize the texture map, we optimize it by minimizing the distance between the rendered view of the texture-projected mesh and the generated patch at each viewpoint using the SGD optimizer. 
Source meshes used for experiments are borrowed from the Objaverse dataset~\cite{objaverse}\footnote{Objaverse: https://huggingface.co/datasets/allenai/objaverse, ODC-By v1.0 License}.
Following prior works~\cite{liu2023text, kim2024synctweedies, lee2025syncsde}, we apply Voronoi diagram-augmented filling~\cite{aurenhammer1991voronoi} and a modified self-attention mechanism in the noise prediction network.
The latent texture map resolution is set to $1536^2$, and the RGB texture map resolution is $1024^2$.
Each control variable $\mathbf{u}_t^{(n)}$ is optimized for three iterations with a learning rate of $10^{-2}$ using the Adam optimizer~\cite{kingma2014adam}. 
We use $\beta = 1.0$, $\gamma = 0.1$, and $\lambda = 2.0$ as default hyperparameter.

For evaluation, we render the textured meshes from 10 different viewpoints using PyTorch3D renderer~\cite{ravi2020accelerating}\footnote{PyTorch3D: https://github.com/facebookresearch/pytorch3d, BSD License}.
Eight views have an elevation of $0^\circ$ and azimuths uniformly sampled from $[0^\circ, 360^\circ)$, while two additional views have an elevation of $30^\circ$ with azimuths of $0^\circ$ and $180^\circ$, which corresponds to front and back view, respectively.
FID~\cite{heusel2017gans} and KID~\cite{binkowski2018demystifying} scores are calculated between the rendered images and the reference sets for each (mesh, prompt) pair, then averaged. 
For each (mesh, prompt) pair, we render depth maps from same 10 viewpoints that are used for evaluation and generate 50 images per each depth map using the pretrained depth-conditioned ControlNet, resulting in 500 reference images. 
CLIP-S~\cite{radford2021learning} is measured by averaging the cosine similarity between each of the 10 rendered views and the corresponding prompt for every (mesh, prompt) pair.
For qualitative visualization in Figure~\ref{fig:mesh_cmp}, we use Nvdiffrast~\cite{Laine2020diffrast}\footnote{Nvdiffrast: https://github.com/NVlabs/nvdiffrast, Nvidia Source Code License (1-Way Commercial)} rasterizer for more sophisticated rasterization.

\paragraph{Additional qualitative results.}

\begin{figure}[h!]
	\centering
    	\includegraphics[width=1.0\linewidth]{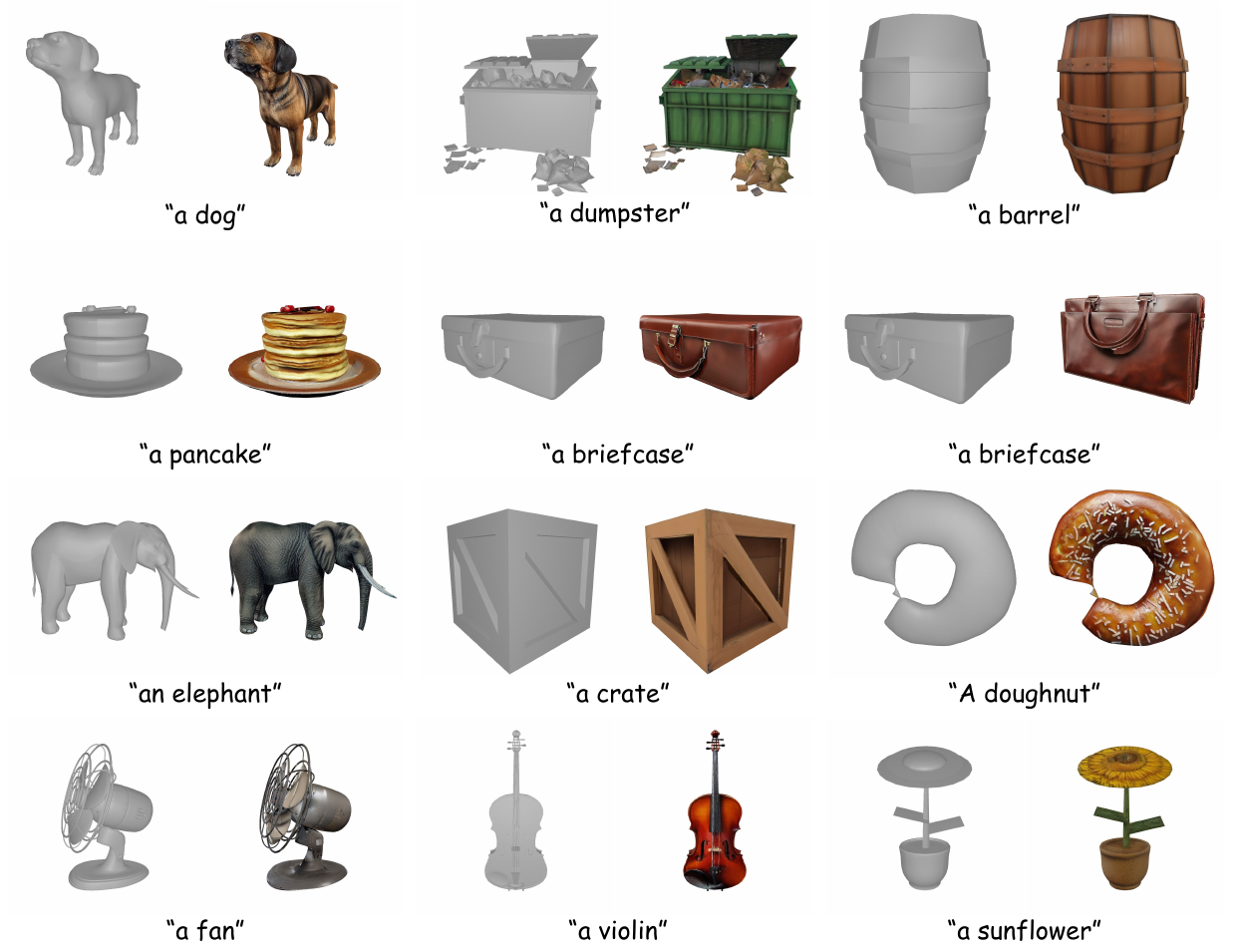}
	\vspace{-6mm}
	\caption{[Best viewed when magnified.] \textbf{Our method generates artifact-free and realistic textures for diverse 3D meshes.}
    }
\label{fig:supp_mesh}
\end{figure}

We show additional qualitative results of \methodname~on text-guided 3D mesh texturing task in Figure~\ref{fig:supp_mesh}.
As shown, our method synthesizes textures that are not only realistic but also rich in fine-grained details.

\clearpage

\subsection{Discussion on computational cost}
\label{sec:supp_computation}

We measure the required runtime of each method to generate a single image for wide image generation and optical illusion generation task.
Runtimes are measured using a single NVIDIA A6000 GPU with the official implementation of each method.
Table~\ref{tab:runtime_wide_image} and~\ref{tab:runtime_illusion} summarize the results.
Since our method involves an optimization while the baselines do not, it incurs a longer runtime. 
Nevertheless, as demonstrated in Table~\ref{tab:wide_image} and~\ref{tab:illusion}, \methodname~achieves stronger performance while maintaining practical usability.

\begin{table}[h!]
\caption{
Quantitative runtime measurement for wide image generation task. 
}
\vspace{-2mm}
\centering
\setlength{\tabcolsep}{2.5pt} 
\scalebox{0.85}{
    \begin{tabular}{l c c c c c}
    \toprule
    Method & MultiDiffusion~\cite{bar2023multidiffusion}  & SyncTweedies~\cite{kim2024synctweedies} & SyncSDE~\cite{lee2025syncsde}  & StochSync~\cite{yeo2025stochsync} & \textbf{\methodname~(Ours)}   \\
    \midrule
    Runtime (s/image) &  68.70    & 73.98  & 17.51 & 42.37 & 232.41 \\
    \bottomrule
    \end{tabular}
    }
    \label{tab:runtime_wide_image}
\end{table} 
\begin{table}[h!]
\caption{
Quantitative runtime measurement for optical illusion generation task. 
}
\vspace{-2mm}
\centering
\setlength{\tabcolsep}{6pt} 
\scalebox{0.9}{
    \begin{tabular}{l c c c c}
    \toprule
    Method & SyncTweedies~\cite{kim2024synctweedies} & SyncSDE~\cite{lee2025syncsde}  & Anagram-MTL~\cite{xu2025diffusion} & \textbf{\methodname~(Ours)}   \\
    \midrule
    Runtime (s/image) & 6.04 & 6.25 & 12.38  & 33.01 \\
    \bottomrule
    \end{tabular}
    }
    \label{tab:runtime_illusion}
\end{table}

\section{Societal impacts}
\label{sec:supp_societal_impacts}

Our method enables collaborative generation in diverse and challenging scenarios, making it applicable to various visual generation tasks that require globally consistent outputs. 
It may improve the practicality of generative models in real-world applications such as content creation and design.
However, it may also inherit the pretrained generative model's potential limitations, including the generation of harmful or unethical content.

\clearpage

\end{document}